\documentclass[twoside,11pt]{article}
\usepackage{jair, rawfonts}

\usepackage{amsfonts}
\usepackage{graphicx}
\usepackage[table]{xcolor} 
\usepackage{dsfont}
\usepackage{tabularx}
\usepackage{mathptmx}
\usepackage{pifont}
\usepackage{booktabs} 
\usepackage{natbib}
\usepackage[colorlinks=true,linkcolor=black,anchorcolor=black,citecolor=black,filecolor=black,menucolor=black,runcolor=black,urlcolor=black]{hyperref}
\usepackage{bm}
\usepackage[parfill]{parskip}

\usepackage{graphicx}
\usepackage{hyperref}
\usepackage{cleveref}
\usepackage{keyval}
\usepackage[ruled,vlined]{algorithm2e}
\usepackage{amssymb}
\usepackage{color}
\usepackage{wrapfig} 
\usepackage{url}
\usepackage[font=small,labelfont=bf]{caption}

\newtheorem{definition}{\textbf{Definition}}

\long\def\/*#1*/{}

\jairheading{74}{2022}{459-515}{01/2022}{05/2022}
\ShortHeadings{Core Challenges in Embodied Vision-Language Planning}
{Francis, Kitamura, Labelle, Lu, Navarro \& Oh}
\firstpageno{459}

\begin{document}

\raggedbottom 
\title{Core Challenges in Embodied Vision-Language Planning}

\author{\name Jonathan Francis \email jmf1@cs.cmu.edu \\
\addr School of Computer Science, Carnegie Mellon University\\ 5000 Forbes Avenue, Pittsburgh, PA, USA
\AND
\name Nariaki Kitamura \email nariaki\_kitamura@global.komatsu \\
\addr Komatsu Ltd.,\\ 2-3-6 Akasaka, Minato-ku, Tokyo, Japan 
\AND
\name Felix Labelle \email flabelle@alumni.cmu.edu \\
\name Xiaopeng Lu \email xiaopen2@alumni.cmu.edu \\
\name Ingrid Navarro \email ingridn@cs.cmu.edu \\
\name Jean Oh \email jeanoh@cmu.edu \\
\addr School of Computer Science, Carnegie Mellon University\\
5000 Forbes Avenue, Pittsburgh, PA, USA}


\maketitle

\newcommand{\cmark}{\ding{51}}%
\newcommand{\xmark}{\ding{55}}%

\newcommand{\jf}[1]{\textcolor{red}{[\textbf{JF:} #1]}}
\newcommand{\fl}[1]{\textcolor{blue}{[\textbf{FL:} #1]}}
\newcommand{\jo}[1]{\textcolor{magenta}{[\textbf{JO:} #1]}}
\newcommand{\na}[1]{\textcolor{cyan}{[\textbf{IN:} #1]}}
\newcommand{\nav}[1]{\textcolor{cyan}{[\textbf{IN:} #1]}}
\newcommand{\xl}[1]{\textcolor{orange}{[\textbf{XL:} #1]}}
\newcommand{\nk}[1]{\textcolor{green}{[\textbf{NK:} #1]}}
\newcommand{\parentcite}[1]{\citep{#1}}%
\newcommand{\parencite}[1]{\citep{#1}}%
\newcommand{\textcite}[1]{\cite{#1}}%
\newcommand{\para}[1]{{\noindent\textbf{#1}}}
\newcommand{\comment}[1]{}
\newcommand{\cut}[1]{}

\newcommand{\ann}[1]{}
\newcommand{\rev}[1]{\textcolor{black}{#1}}
\newcommand{\upd}[1]{\textcolor{black}{#1}}

\newcommand{\EVLP}{Embodied Vision-Language Planning}
\newcommand{\EMP}{Embodied Multimodal Planning}
\newcommand{\evlp}{EVLP}
\newcommand{\emp}{EMP}

\begin{abstract}

Recent advances in the areas of multimodal machine learning and artificial intelligence (AI) have led to the development of challenging tasks at the intersection of Computer Vision, Natural Language Processing, and Embodied AI. Whereas many approaches and previous survey pursuits have characterised one or two of these dimensions, there has not been a holistic analysis at the center of all three. Moreover, even when combinations of these topics are considered, more focus is placed on describing, e.g., current architectural methods, as opposed to \textit{also} illustrating high-level challenges and opportunities for the field. In this survey paper, we discuss \EVLP~(\evlp) tasks, a family of prominent embodied navigation and manipulation problems that jointly use computer vision and natural language. We propose a taxonomy to unify these tasks and provide an in-depth analysis and comparison of the new and current algorithmic approaches, metrics, simulated environments, as well as the datasets used for \evlp~tasks. Finally, we present the core challenges that we believe new \evlp~works should seek to address, and we advocate for task construction that enables model generalizability and furthers real-world deployment. 

\end{abstract}

\section{Introduction}\label{sec:intro}

With recent progress in the fields of artificial intelligence (AI) and robotics, intelligent agents are envisaged to cooperate or collaborate with humans in shared environments. Such agents are expected to understand semantic contexts of an environment, e.g., using visual information perceived using sensors, as well as auditory or textual information intended for humans, e.g., presented in natural language. With the goal of developing intelligent agents disposing of these capabilties, embodied AI is a field that studies AI problems situated in a physical environment. Recently, the number of papers and datasets for the tasks that require the agents to use both vision and language understanding has increased markedly~\citep{EQA,IQA,R2R,VLN-CE,CVDN, nguyen2019hanna,VLNBERT,FineGrainedR2R}. In this article, we conduct a survey of recent works on these types of problems, which we refer to as as \textit{\EVLP}~(\evlp)~tasks. In this article, we aim to provide a bird's-eye view of current research on \evlp~problems, addressing their main challenges and future directions. \textbf{The main contributions of this article are the following}:
\begin{enumerate}
    \item We propose a taxonomy that unifies a set of related subtasks of \evlp~
    \item We survey the current state-of-the-art techniques used in the \evlp~family to provide a roadmap for the researchers in the field
    \item More importantly, we identify and discuss the remaining challenges with an emphasis towards solving real-world problems. 
\end{enumerate}

\subsection{Scope of this Survey}\label{ssec:scope}
An ``agent" in this article refers to an entity that can make decisions and take actions autonomously \parencite{wooldridge1995intelligent, castelfranchi1998modelling}. An embodied agent is situated in a physical or virtual environment which the agent navigates in and interacts with. An embodied agent is generally equipped with sensing capabilities, e.g., via visual or auditory sensing modalities. In this article, we focus on existing works that use visual perception and language understanding as the two major inputs for an embodied agent to make decisions for various tasks in its environment. We note that the topics covered here are mainly for a single agent in a static environment, i.e., \ann{Reviewer C, ``Claims", item \#1} \rev{topics on multi-agent planning are outside the scope of this article}. \ann{Reviewer C, ``Claims", item \#2} \rev{Moreover, as the majority of tasks discussed in this article are conducted through simulated environments, real-world physical challenges in robotics (e.g., visual affordance learning, proprioceptive control, system identification) are not considered.} \upd{\evlp~problems have previously been studied in the fields of natural language processing, robotics, and computer vision \parencite{TowardsReasoning, boularias2015grounding, duvallet2016inferring}. While the focus of this survey is on contemporary works, we will discuss how concepts from classical approaches have inspired recent methodology and how they could be used for future directions, e.g., as in the use of mapping and exploration strategies, search and topological planning, and hierarchical task decomposition (Section \ref{ssec:navigation}).}

\subsection{Intended Audience and Reading Guide}
This paper is tailored to accommodate a broad spectrum of reader backgrounds and perspectives, as this paper is positioned at the intersection of Computer Vision (CV), Natural Language Processing (NLP), and Robotics. For readers that are new to these topics, we provide an in-depth coverage of existing works and methodologies; for readers with significant experience in one or more of these areas, we offer a taxonomy of the broader field of \textit{\EVLP}~in~\Cref{ssec:taxonomy}, provide an analysis of core challenges, and discuss future directions in~\Cref{sec:openchallenges}. 

For readers who are less familiar with implementation and evaluation of embodied agents, it is recommended to read the sections in order, as each section builds upon previous sections. The rest of the article is organized as follows. In~\Cref{sec:prob-def}, we formally define the class of \evlp~problems discussed in this article and present a taxonomy of the field. After describing the key tasks that compose the \evlp~family in~\Cref{sec:tasks}, we present modeling approaches including commonly-used learning paradigms, architectural design choices, and techniques proposed to tackle \evlp-specific challenges. \Cref{sec:evaluation} presents the datasets and evaluation metrics currently used by the research community. Finally, \Cref{subsec:challenges} discusses several open challenges in the field.

\subsection{Related Surveys}\label{ssec:related-work}
We review existing survey articles on relevant topics, in order to provide readers with pointers to other papers on more specific topics and to clarify how this article differs from them.

\begin{itemize}
    \item \textcite{IndoorEmbodiedAgentSurvey} discuss the datasets, methodologies, and common approaches to Vision Indoor Navigation (VIN) tasks. VIN tasks do not necessarily require language, are limited to navigation, and only occur in indoor environments. Our survey touches on planning tasks, of which navigation is a subset, which use language to establish the goal state. Also, \evlp~tasks are not limited to indoor environments, but also include outdoor settings. 
    
    \item \textcite{RLLanguageSurvey} focus on existing approaches of using language components in the context of reinforcement learning (RL). The authors divide the language-related problem into language-conditional RL and language-assisted RL. In contrast, our work briefly discusses different training paradigms commonly used for tasks in the~\evlp~domain, in \Cref{sec:approaches}.
    
     
    \item \textcite{MultiModalSurvey} cover popular tasks and approaches in multimodal research. Our survey, in comparison, specifically focuses on \evlp~tasks, and provides more details, such as commonly used datasets, training paradigms, and challenges on this specific domain.
    \item \ann{Reviewer C, ``Related Works", item \#1} \rev{\textcite{bisk2020experience} explore the current research in language understanding and its shortcomings. To categorize the differences between human and machine understanding, the authors define the notion of a ``world scope.'' A worldscope defines the type of real world information that language models capture, such as physical relationships or social concepts. Five worldscopes are used to define a framework that discusses what aspects of language are harder to grasp and issues with the current approaches. We do not set out a general framework to identify the complexity of language. Instead, we survey the low-level research progress of ~\evlp~tasks, discuss the role of language in planning tasks, and further point out possible improvements.}
    
    \item \textcite{tangiuchi2019survey} discuss techniques at the intersection of language and robotics research. Their work focuses heavily on the syntactic and semantic structures of language information, and how it can be tied to the low level robotics actions. Instead, our survey focuses primarily on the task level, as opposed to low-level actions, and is suitable for readers with no prior knowledge in the robotics field.
    
    \item \textcite{uppal2020emerging} provide discussion of Vision and Language problems, dividing Vision-language tasks into four categories: generation, classification, retrieval, and other tasks. Here, the authors only discuss one \evlp~task; however, due to the increased popularity of embodied vision-language planning, we feel that the \evlp~family warrants its own specific treatment and careful discussion. Moreover, \textcite{uppal2020emerging} further discuss changes in representation such as the use of transformers, fusion of multiple modalities, architectures, and evaluation metrics. While both surveys discuss changes in multimodal representation, ours focuses on \evlp~research, providing a unified problem definition, taxonomy, and analysis of trends for this exciting field.

    \item \cite{batra2020rearrangement} provide a detailed analysis and framework for the rearrangement task, which involves using an embodied agent to change the environment from an initial state to a target state. The authors argue that the manipulation problem can be viewed as two-fold: agent-centric problem and environment-centric problem. They further analyze the more fine-grained split for each problem. Several evaluation metrics and existing testbeds are presented. While they do introduce a framework which includes one \evlp~task,  it is not the principle focus of that paper. We pursue a broader discussion of emerging trends and core challenges in the field. 
\end{itemize}

\section{Problem Definition}\label{sec:prob-def}

In this article, we discuss a broad set of problems, related to an embodied agent’s ability to make planning decisions in physical environments. 
We include both stateless problems such as question answering and sequential decision-making problems, such as navigation, under the umbrella of ``planning''. Note that stateless problems belong to the special case of single-step decision making. More generally, we refer to sequential decision-making problems as those where an agent must perform actions over a countable time-horizon, given an objective. In the context of navigation, for example, an agent may choose actions, according to a pre-defined action space, that enable an agent to transition from an initial state to a goal state. In the context of manipulation, the agent executes a series of actions, in order to effect a desired change in the environment. More formally, planning problem are defined as follows:

\begin{definition}[Planning]  Let $S$ denote a set of states; $A$, a set of actions. A planning problem is defined as a tuple of its states, actions, initial and goal states: $\Phi = \{S, A, s_{ini}, s_{goal}\}$, where $s_{ini}, s_{goal} \in S$ denote initial and goal states, respectively. 
A solution $\psi \in \Psi_{\Phi}$ to planning problem $\Phi$ is a sequence of actions to take in each state, starting from an initial state to reach a goal state, $\psi = [s_{ini}, a_1, ..., a_T, s_{goal}]$, where $T$ is a finite time-step and $\Psi_{\Phi}$ is a set of possible solutions to $\Phi$.
\end{definition}

\evlp~problems require planning in partially-observable environments, i.e., the entire state space may not be known to the agent in advance. Instead, an agent needs to use vision and language inputs to estimate its current and goal states in order to accomplish high-level tasks in a physical environment. These inputs can be given at the start of a task, e.g., a task itself is described in natural language in the form of a question or instruction, or become available as an agent moves through the state space. Visual inputs are an example of such an input as they are primarily online information that can be perceived through sensing. Note this is not always the case, as they can also be provided as part of a task specification, e.g., a view from a goal location.

\begin{definition}[Embodied Vision-Language Planning (\evlp)]

\ann{Reviewer B, item \#1 (``Definition 2'')} Let $V$ and $L$ denote sets of vision and language inputs available to an agent. Given an \evlp~problem $\Phi$, state $s_t$ at time step $t$ can be defined in terms of vision and language inputs up to the current time step, such that, $s_t = f(v_{1},\dots,v_{t}, l_{1},\dots,l_{t})$ where $v_{t} \in V$ and $l_{t} \in L$. The objective here is to minimize the difference between an \rev{admissible}\footnote{\rev{Admissibility condition: a solution that is possible, under the constraints of the environment's transition dynamics as well as the agents' own system dynamics, which satisfies the task specification.}} solution $\psi \in \Psi_{\Phi}$ and a predicted one $\bar{\psi}$. 
\end{definition}

This definition broadly captures the crux of \evlp~problems. A customized definition would be needed for each specific task where additional constraints or assumptions are added to focus on particular subareas of this general problem. For instance, Vision-Language Navigation (VLN) is a natural language direction following problem in an unknown environment, which can be defined as a planning problem where an agent is given an initial state and a solution (or a sequence of actions) represented in natural language, and is equipped with visual perception, e.g., first-person view images.

\subsection{Taxonomy}
\label{ssec:taxonomy}

\begin{figure}[!tp]
    \centering
    \includegraphics[width=\textwidth,keepaspectratio]{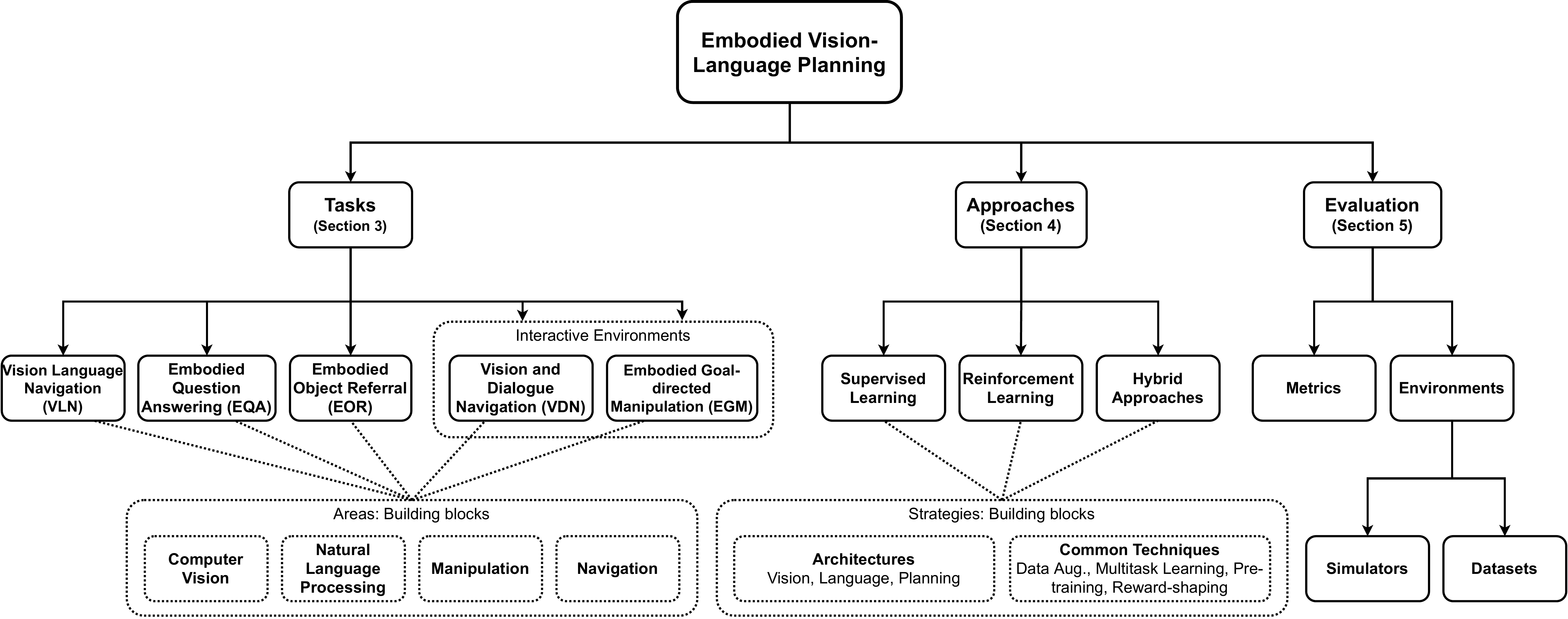}
    \caption{\small Taxonomy of \textit{\EVLP}, aligned with our paper organisation.} 
    \label{fig:taxonomy}
\end{figure}

We propose a taxonomy of current \evlp~research, illustrated in \autoref{fig:taxonomy}, around which the rest of the paper is organized. The taxonomy subdivides the field into three branches; tasks, approaches, and evaluation methods. The Tasks branch (corresponding to \hyperref[sec:tasks]{Section \ref{sec:prob-def}}) proposes a framework to classify existing tasks and to serve as a basis for distinguishing new ones.

The Approaches branch (\hyperref[sec:approaches]{Section \ref{sec:approaches}}) touches on the learning paradigms, common architectures used for the different tasks, as well as common tricks used to improve performance. The technical challenges that underlie all tasks are discussed in \hyperref[sec:common-tech]{Section \ref{sec:common-tech}}.

The right-most branch of the taxonomy, in Figure \ref{fig:taxonomy}, discusses task Evaluation Methodology (\hyperref[sec:evaluation]{Section \ref{sec:evaluation}}), which is subdivided into two parts: metrics and environments. The metrics subsection references many of the common metrics and their formul\ae, used throughout \evlp~tasks, while the environments subsection presents the different simulators and datasets currently used.

\subsection{Tasks in \EVLP} 
\label{sec:tasks}

\ann{Reviewer B, item \#2 (``Missing citations")} Many \evlp~tasks have been proposed, with each task focusing on different technical challenges and reasoning requirements for agents. Tasks vary on the basis of the action space (types and number of actions possible), the reasoning modes required (e.g., instruction-following, versus exploration and information-gathering), and whether or not the task requires dialogue with another agent. In this survey, we only include tasks where clearly-defined datasets and challenges exist for evaluation benchmarking, i.e., Vision Language Navigation (VLN) \parencite{R2R,misra2018mapping,StreetNav,R4R,ku2020roomacrossroom}, Vision and Dialogue History Navigation (VDN) \parencite{devries2018talk,nguyen2019hanna,CVDN}, \rev{Embodied Question Answering (EQA) \parencite{EQA,eqa_matterport}}, Embodied Object Referral (EOR) \parencite{qi2020reverie,TouchDown}, and Embodied Goal-directed Manipulation (EGM) \parencite{ALFRED,ArraMon,CerealBar}. In the following sub-sections, we compare and contrast these task families, with the hope that readers gain a solid understanding of the differences in the problem formulations. In this manner, new datasets can be contextualized by existing tasks, and new task definitions can be later positioned alongside those mentioned here. We summarize the attributes for existing tasks in~\Cref{tab:CompVLN}. We further refer the readers to to \Cref{ssec:datasets} for more information about the datasets and simulation environments pertaining each task.

\begin{table}[!htp]
\footnotesize
\caption{\small \ann{Reviewer B, item \#1 (``Table 1 caption'')} \ann{Reviewer C, ``Minor Comments'', item \#1} Properties of Embodied Vision-Language Planning (EVLP) tasks: `Navigation' describes whether navigation is part of the actions for the task; \rev{`Object Identification' indicates whether the agent may only solve the task through identifying specified objects}; \rev{`Environment Interaction' indicates whether or not the agent can mutate environment state, e.g., through dialogue-based interaction with another agent or through object manipulation}; and `Primary Reasoning Mode(s)' describes how agents are intended to interpret the task.}
\label{tab:CompVLN}
\begin{tabularx}{\textwidth}{c|lllX}
 Task & Navigation & \rev{Object Identification} & \rev{Environment Interaction} & \upd{Primary Reasoning Mode(s)}\\
 \hline
VLN & \cmark & \xmark  & \xmark & \upd{Understanding object and scene layout properties; instruction-following} \\
EQA & \cmark & \cmark  & \xmark & \upd{Exploration and information-gathering} \\
EOR & \cmark & \cmark & \xmark &  \upd{Understanding spatio-semantic object relations and scene layout properties; instruction-following}\\
VDN & \cmark & \xmark  & \cmark  & \upd{Understanding object and scene layout properties; multiple instruction-following} \\ 
EGM & \cmark & \cmark & \cmark & \upd{Understanding object affordances, environment attributes, and scene layout properties; instruction-following}\\
\end{tabularx}
\end{table}

\subsubsection{Vision Language Navigation}
\label{subsubsec:vln}

Vision-Language Navigation (VLN) requires an agent to navigate to a goal location in an environment following an instruction $L$. A problem in VLN can be formulated as $\Phi_{VLN}=\{S, A, s_1, s_{goal}\}$, where a solution or a path $\psi_{\Phi_{VLN}} = \{s_1, a_1, \dots, a_{T}, s_{goal}\}$ exists, such that each state $s_{t} \in S, t \in [1, T]$ is associated with a physical location in the environment leading to the target location. The action space $A$ available to the agent consists of navigation actions between physical states, and a \texttt{stop} action which determines the end of a solution. Navigation actions can be discrete, e.g., \texttt{turn\_left}, \texttt{turn\_right} or \texttt{move\_forward} \parencite{R2R, VLN-CE}, as well as continuous \parencite{RoboVLN}. Finally, the goal of the agent is to predict a solution $\bar{\psi}_{\Phi_{VLN}}$ consisting of a sequence of actions $a_t \in A, t \in [1, T]$ that closest align to the instruction, and thus, to the true solution $\psi_{\Phi_{VLN}}$.

VLN is the most established \evlp~task: a number of datasets (See Section~\ref{ssec:datasets}) exist in both, indoor \parencite{R2R, R4R, VLN-CE, RoboVLN}, and outdoor environments \citep{StreetNav, misra2018mapping}. Overall, VLN models have seen considerable progress in improving the ability to get closer to the goal and to the ground truth trajectory \parencite{fried2018speaker, tan2019learning, FineGrainedR2R, VLNBERT, R4R}. Nonetheless, \textcite{zhu2021diagnosingVLN} show that it is unclear if models are actually aligning the visual modality and that recent work has experienced a slow-down in performance improvements. They suggest that understanding how VLN agents interpret visual and textual inputs when making navigation decisions is an open challenge.

\subsubsection{Embodied Question Answering}
\label{subsubsec:eqa}

In Embodied Question-Answering (EQA), an agent initially receives a language-based question $L$, and must navigate around collecting information about its surroundings to generate an answer. An EQA problem can be formulated as $\Phi_{EQA} =\{S, A, R, s_1, r\}$, where $R$ is the set of possible answers, and $r \in R$ represents the correct answer to the given question. The actions $A$ available to the agent consist of discrete moving actions as those in Subsection~\ref{subsubsec:vln}. However, as opposed to using a \texttt{stop} action, the ending of an episode is given by an \texttt{answer} action \parencite{EQA, IQA}. Lastly, a correct solution to a given question can be expressed as $\psi_{\Phi_{EQA}} =\{s_1, a_1,\dots,a_{T}, r\}$. 

Unlike the aforementioned tasks, there are some challenges unique to EQA. First, there might not necessarily exist a unique or perfect solution for any given question. Second, EQA also requires an agent to understand the implications of the given questions and to translate them into actions that lead to the correct answer. Furthermore, the EQA task requires an agent's awareness of its environment and commonsense knowledge on how those environments function or are spatially organized, e.g., food is often located in the fridge, or parked cars are often found in a garage. The level of ambiguity, reasoning required, and knowledge required make EQA one of the more challenging EVLP tasks to date.

\subsubsection{Embodied Object Referral}
\label{subsubsec:eor}

In Embodied Object Referral (EOR) tasks, an agent navigates to an object $o$ mentioned in a given instruction $L$, and has to identify (or select) it upon reaching its location. EOR can be framed as $\Phi_{EOR} =\{S, A, O, s_1, s_{goal}, o\}$. Here, $O$ is the set of possible objects in the environment, which are specified by a class label and a bounding-box or a mask, and $o \in O$ is the object of interest. The set of available actions $A$ includes discrete navigation actions as those described in Subsection~\ref{subsubsec:vln}. An EOR solution can be expressed as $\psi_{\Phi_{EOR}} =\{s_1, a_1,\dots,a_{T}, s_{goal}, o\}$, and represents the path leading to the object of interest $o$. When the final state $s_{goal}$ is reached, a bounding box or a mask is used to indicate where the object $o$ is located in the viewpoint given at that state. Finally, the goal of the agent is to predict a solution $\bar{\psi}_{\Phi_{EOR}}$ that follows the provided instruction and correctly selects the referenced object, and thus, closely matches solution $\psi_{\Phi_{EOR}}$.

Similar to VLN, EOR relies on instruction following. In this setting, instructions can be step-by-step \parencite{TouchDown, mehta2020retouchdown}, or they can be under specified \parencite{qi2020reverie}. The latter requires the agent to reason and understand the context of the instruction, as was discussed in Subsection~\ref{subsubsec:eqa}. In addition, EOR introduces the challenge of identifying an object of interest. This object might be visible from different viewpoints, and like previous tasks, multiple instances of such object can exist in the environment. More importantly, while an object might be explicit \parencite{TouchDown}, an agent might also be required to predict what object it should find from a given instruction \parencite{qi2020reverie}. 

\subsubsection{Vision and Dialogue Navigation}
\label{subsubsec:vdn}

A language instruction can be ambiguous and might require clarification or assistance \parencite{CVDN, nguyen2019vision}. For example, when navigating through a home, an agent might find multiple instances of an object referenced in an instruction. Moreover, an instruction may not specify the goal in the level of granularity that the agent needs to plan and execute actions; the agent may require clarification on intermediate sub-tasks. Unlike VLN and EQA, VDN allows an agent to interact with another agent (e.g., a human collaborator) to resolve these types of uncertainties. This interactive element has been added in two ways: through the use of an ambiguity resolution module \parencite{JustAsk, nguyen2019hanna}, or through dialogue \parencite{CVDN, nguyen2019vision}. In the case of the former, if the agent gets confused, it may ask for help from an oracle who is aware of the agent's state and its goal. In the latter, the agent is given an initial vague prompt and is required to ask an oracle for clarification. 

More formally, Vision and Dialogue Navigation (VDN) requires an agent to navigate to a goal location in an environment, but may do so through sequential directives $L = \{l_1, \dots, l_N\}$. These directives may come in the form of sub-instructions upon an agent's request, or as dialogue-based interaction. Here, each instruction represents a sub-problem $\Phi_{VDN}^{(i)} =\{S, A, s_1^{(i)}, s^{(i)}_{goal}\}_{i=1}^N$, leading the agent closer to its ultimate goal position. The actions $A$ that the agent can execute consist of navigation actions, as in previous tasks, but may additionally include other forms of interactive actions (e.g., to request help). Akin to VLN, a solution to a VDN problem can be denoted as the set $\psi^{(i)}_{\Phi_{VDN}} = \{s_1^{(i)}, a_1^{(i)}, \dots, a_{T}^{(i)}, s_{goal}^{(i)}\}_{i=1}^{N}$, and the goal of the agent is to find $\bar{\psi}_{\Phi_{VDN}}^{(i)}, i \in [1, N]$, a feasible set of solutions that best align with the true solutions.

\upd{We note that past works in VDN \parencite{CVDN, nguyen2019vision} use static, multi-turn question-answer pairs to simulate conversation between the ego-agent and another agent. Still, we accommodate, as part of the VDN family, problem definitions that feature active dialogue context as well.} 

\subsubsection{Embodied Goal-Directed Manipulation}
\label{subsubsec:egm}

\ann{Reviewer C, ``Taxonomy'', item \#1} \rev{Unlike previous tasks, Embodied Goal-directed Manipulation (EGM) requires the manipulation of objects in a scene, posing unique challenges for agents \parencite{ALFRED, padmakumar2021teach}. EGM may combine these manipulation-based environment interactions with requirements from aforementioned tasks, such as navigation and path-planning, state-tracking, instruction-following, instruction decomposition, and object selection \parencite{ArraMon}. Due to shared properties between EGM and, e.g., VLN task definitions, EGM also encompasses the mobile manipulation paradigm from previous literature \parencite{tellex2011understanding, khatib1999mobile}.}

EGM may require solving multi-step instructions and, as such, it can be framed as a set of multiple sub-problems $\Phi^{(i)}_{EGM} =\{S, A, O, s_1^{(i)}, s^{(i)}_{goal}\}_{i=1}^N$ where the $i_{th}$ sub-problem is associated to the $i_{th}$ sub-instruction. In this setting, the action space $A$ of the agent includes navigation actions as in the previous tasks, in addition to actions involving interactions with objects. These interactions can be sub-divided into multiple types of interactions, e.g. \texttt{pick\_up}, \texttt{turn\_on}, \texttt{turn\_off}, etc. \parencite{ALFRED}, or used as a single \texttt{interact} action \parencite{ArraMon}. Then, $\psi_{\Phi_{EGM}}^{(i)} = \{s^{(i)}_1, a^{(i)}_1,\dots,a^{(i)}_{T}, s^{(i)}_{goal}\}_{i=1}^{N}$ represents a solution to an EGM problem. Here, a state $s_t$ not only includes the visual observation of the agent's location at time step $t$, but also the state of objects that the agent interacted with. Lastly, the goal of the agent is to predict $\bar{\psi}^{(i)}_{\Phi_{EGM}}, i \in [1, N]$ that follows the provided instruction(s) and correctly selects and interacts with the object(s) referred in it, and thus, best matches the true solution(s).

In EGM, there may be constraints such as which objects can or can't be picked up. To interact with an object, the agent specifies where $o$ is in its view using a bounding box or mask. Furthermore, agents not only need to interpret instructions and recognize objects they are meant to interact with, but also need to understand the consequences of interacting with their environment. Immutability is an important constraint, in real life you cannot un-slice a tomato. This means certain mistakes lead to ``un-winnable" states and that there may be an order of operations for certain tasks. This makes EGM one of the most difficult \evlp~task.

\section{Approaches}
\label{sec:approaches}

This section provides a review of technical approaches used in \evlp. It discusses how current works tackle the different facets of \evlp~tasks, namely vision, language, and planning. It then discusses different learning paradigms and  tricks used to improve performance.

\subsection{Modeling Vision, Language, and Planning}

\subsubsection{Modeling Vision}\label{subsubsec:cnns}

Modeling vision is an important part of \evlp~tasks, as it is the principle way in which agents build a representation of their environment. Representations of vision used include the use of explicit features \parencite{DuvalletThesis,tellex2011understanding,duvallet2016inferring} and neural representations. In \evlp~models, neural representations are most common, with Convolutional Neural Networks (CNNs) being used as encoders in most works~\parencite{MultiModalSurvey}. 
CNNs are neural network architectures suitable to process structured data such as images, and thus are broadly used in the computer vision tasks, e.g., image classification~\parencite{CNNsforImageClassification}. 

In existing works, pre-trained neural networks such as ResNet~\parencite{Resnet} are commonly used for extracting meaningful image features. One limitation of using such pre-trained networks is that these features may lead to over-fitting. \citet{EnvBias} propose using the logits of the classification layer--i.e., a higher-level feature--in order to reduce the performance gap between the seen and the unseen environments. 

Neural feature approaches perform well, but lack the interpretability of using explicit features. \citet{Anderson2017up-down} proposed a hybrid approach, extracting salient features using a pretrained object detector \parencite{maskrcnn} and encoding them using a pretrained CNN. \citet{VLNBERT} use this technique in VLN to represent objects in an image rather than the entire image. This lends well to the use of multimodal transformers, which we will discuss in section \ref{ssec:multimodal}.

\subsubsection{Modeling Language}
\label{subsubsec:rnns}

Goal understanding is another critical piece of the \evlp~puzzle. Goal information, often in the form of instructions or a statement, are provided in the form of language. Like vision, hand-selected features \parencite{DuvalletThesis,tellex2011understanding} or neural representations can be used. Modeling language is primarily done using Recurrent Neural Networks (RNNs) or Transformers.

RNNs are a commonly used neural network architecture for processing sequential data such as language. For more information on RNNs we refer readers to \cite{lipton2015critical}. When used as a language encoder RNNs take in tokenized instructions and generates a vectorized representation of language. RNNs have some short comings, for example issues with long-term dependencies. To better handle long-term dependency, gated RNNs such as Long Short-Term Memory (LSTMs)~\parencite{LSTM} and Gated Recurrent Unit (GRUs)~\parencite{cho2014learning} are widely used for general NLP tasks as well as \evlp~ \parencite{EQA,VLN-CE, fried2018speaker, nguyen2019hanna, CVDN, tan2019learning,EnvBias} due to their superiority in handling long-term dependencies when compared to vanilla RNNs. Other approaches to tackle this include the use of attention, which allows models to attend over the entire context or memory modules \parencite{CrossModalMemory}. 

Transformers operate on a different principle than RNNs, using stacked attention to combine information from different inputs \cite{vaswani2017attention}. Transformer-based architectures have shown to be effective across various communities, such as CV \parencite{khan2021transformers}, NLP \parencite{vaswani2017attention,devlin2018bert}, robotics \parencite{Fang_2019_CVPR, dasari2020transformers}, and multimodal tasks \parencite{shin2021perspectives}. Earlier work has leveraged transformers as language models to improve performance on downstream tasks \parencite{devlin2018bert}. Several works in \evlp~ have used transformers. \textcite{li2019robust} use a BERT encoder to encode language for VLN task which improves upon past models. 

\subsubsection{Modeling Multimodality}
\label{ssec:multimodal}

Representing language and vision can not be done independently from one another. First, there is overlapping information between the two  modalities which could help correct for errors in one representation. Moreover, information from each is required to interact with one other for the overall task, for example grounding landmarks from the text to the environment. In contemporary works, combining modalities is done either through the use of attention or transformers.

Attention was originally proposed to improve the performance of Seq2Seq neural machine translation \parencite{bahdanau2016neural, MTSurvey}. It primarily serves two purposes in \evlp: to fuse modalities; to align modalities \parencite{MMMLTaxonomy}. Attention mechanisms come in many forms, but typically can be framed as a process which takes in two inputs, referred to as the key and query, and produces an output using a weighted sum of the input, known as the value \parencite{vaswani2017attention}. Attention weights the query using the key, which can originate from the same or different sources. In the multimodal context this allows the model to combine vision, language, and the agent's current state \parencite{wang2019reinforced, tan2019learning, fried2018speaker, EQA, qi2020reverie}. This is often in done with the use of multiple attention mechanisms \cite{qi2020reverie,fried2018speaker,tan2019learning}.

More recently, multimodal transformers have also shown to be effective for vision-language tasks \parencite{LXMERT,lu2019vilbert}. For example, \textcite{lu202012} learn language representations together with image information over 12 different multimodal tasks, outperforming most of the tasks when trained independently. In VLN two works have used transformers to combine different modalities. \textcite{PREVALENT} pre-train on instruction, image, action triplets across different tasks and then finetune on a target task. \textcite{VLNBERT} use  VilBERT~\parencite{lu2019vilbert} and pre-train over the conceptual captions dataset \parencite{ConceptualCaptions} to learn multimodal representations using a masking objective for both text and visual streams. During the task-specific fine-tuning, they fine-tune on the VLN task and that shows significant improvement. 
\upd{\subsubsection{Modeling Action-Generation and Planning}}
\label{ssec:navigation}
\upd{At a high level, we measure agents' understanding of a given goal, its scene context, and its understanding of how to \textit{satisfy} goals by its ability to generate appropriate sequences of actions to execute in the environment. In fully-observable (known) environments, where the location of the goal position is known (whether in a global or local/relative coordinate system), we expect the agent to generate a sequence of actions for reaching the goal, with allowance for re-planning at arbitrary frequency, based on newly-acquired information. In partially-observable (partially-known) environments, e.g., where the agent only has access to information about the adjacent admissible states, the agent must generate actions over significantly shorter horizons, such as one or two time-steps. Inspired by classical motion-planning techniques in robotics, several planning approaches have been employed in \evlp~tasks, such as: mapping and exploration, search and topological planning, and hierarchical task decomposition.} 
\upd{\paragraph{Mapping and exploration strategies.}
In the robotics literature, mapping is a general concept which refers to a transformation of the agent's observations into a more abstract state representation, wherein, e.g., planning can be performed more efficiently \parencite{filliat2003map}. Multiple choices of map representation are possible (metric maps, as in occupancy grids, or topological graphs), depending on the required level of expressivity for agent pose, obstacle locations, and goal position \parencite{cummins2008fab}.} 

\upd{Early mapping strategies relied simply on exploring unknown environments. Here the common objective is to prioritise visiting unmapped states at the extent of the currently-explored regions (frontier nodes), in order to facilitate quick and efficient coverage of the environment \cite{burgard2005coordinated}. Agents must estimate the cost of visiting each frontier node and assess whether to proceed in its current exploration heading, versus back-tracking to other frontier nodes that promise more information at lower cost. \textcite{FAST} propose adding an exploration module to VLN agents, allowing agents to perform local decision-making while utilising global information to back-track when the agent gets confused (i.e., revisits previously-visited states). \textcite{ProgressMonitor, RegretfulAgent} propose the \textit{Regretful Agent} (RA), which adds two differentiable modules to the standard VLN architecture: the first, \textit{progress marker} (PM), estimates the agent's progress towards the goal, while the second module, \textit{regret module} (RM), decides whether the agent should back-track by comparing the agent's current observation with its historical information. This module attends over both image and language inputs, and if the weight on the previous image features is larger it will back-track.}

\upd{Because the overhead from mapping an unseen (unknown) or partially-observable environment can be significant, classical robotics approaches moved quickly to Simultaneous Localization and Mapping (SLAM) techniques \parencite{thrun1998probabilistic, cummins2008fab}, wherein an agent projects its observations to a map that it maintains, while also tracking its location in the map. This serves as a basis for registering objects to specific locations of the map and for generating more efficient plans. \cite{ChasingGhosts} use a navigation approach similar to SLAM, which consists of three modules: a mapper, a Bayesian filter, and a policy. The mapper generates a semantic map and the Bayesian filter uses RGB images and depth to find the most likely path at each time step. The agent builds up its understanding of the world and assigns probabilities to different locations. By mapping, an agent can navigate efficiently by avoiding redundant visitations and eventually plan using global information as opposed to local.}
\upd{\paragraph{Search and topological planning.} A popular planning approach is to utilise search algorithms, such as informed search \parencite{hart1968formal, stentz1995focussed}, commonly employed for traversing a graphical state representation (e.g., a tree or topological map). Greedy search algorithms \parencite{black1998dictionary}, wherein the most likely action is taken at each state, is of particular interest for partially-observable environments, where the agent has limited information to use for deciding on the next state transition.} 

\upd{In the context of \evlp~problems, it is common to implement greedy search by way of neural sequence-to-sequence (Seq2Seq) modeling---serving as one of the earliest action-generation strategies reported for these tasks \parencite{fried2018speaker, EQA, R2R, ALFRED}. One shortcoming of this approach is that it does not account for the likelihood of an entire sequence, meaning that it can generate sub-optimal output sequences. Beam search \parencite{holtzman2019curious} is an alternative, which accounts for this by looking at $k$ possible output sequences. Other search-inspired sampling algorithms, such as top-k sampling or nucleus sampling, have shown additional promise, primarily in natural language processing community \parencite{holtzman2019curious}.}

\upd{Other approaches couple mapping with greedy search strategies. \textcite{EvolvingGraphicalPlanner} propose the \textit{Evolving Graphical Planner}, which creates a graph of all admissible nodes (states), where each node is represented using learnable embeddings of the agent's scene context. The agent's global planning module attempts to generate actions based on the current version of the map: the map, trajectory history, and language components are fed into the planner model, which returns a probability distribution over the set of the next node candidates; the agent then samples the next location and builds up a new graph based on the new information at this node, keeping only the top-$K$ locations at each time step.}
\upd{\paragraph{Hierarchical task decomposition.} Early robot planning approaches considered multiple levels of abstraction, defined for an arbitrary planning task: the highest level of abstraction need only contain the goal specification, a lower level of abstraction would contain a sequence of subgoals needed to satisfy the task and reach the goal, and even lower abstraction levels would include the primitive actions required for satisfying each sub-goal \parencite{sacerdoti1974planning, nau2003shop2}. An agent is said to perform hierarchical task decomposition if it is capable of reasoning at multiple levels of task abstraction, as a consequence of either formal predicate calculus \parencite{nilsson1984shakey, kaelbling2010hierarchical} and/or simply based on the agent's architectural specification.}

\upd{Architecturally, it is common to define at least two modules, e.g., a global planner and a local planner, where the former is responsible for generating intermediate sub-tasks towards a given goal, and the latter is responsible for generating satisfactory low-level actions for each sub-task. These ideas manifest in more recent works as, e.g., hierarchical waypoint prediction and navigation \parencite{chen2020learning, misra2018mapping, blukis2018mapping} and hierarchical reinforcement learning \parencite{li2020hrl4in, nachum2018data}. Certain datasets \parencite{misra2018mapping, ALFRED, CerealBar, RoboVLN} pair sub-goal sequences with a high-level goal, allowing for hierarchical approaches to learn to reason at multiple levels of abstraction: agents can predict abstracted actions, corresponding to sub-goals (e.g., a waypoint to navigate to), and use a lower-level module to control the movement actions to achieve that sub-goal (e.g., motor movements to transition between waypoints). \textcite{misra2018mapping, blukis2018mapping} propose a two-part modular architecture: the first is a Visitation Distribution Prediction (VDP) module, which predicts waypoints using RGB images and instructions on a probabilistic map; and the second stage receives this map from the VDP module, for learning to predict low-level actions, through reinforcement learning.}

\ann{Reviewer B, item \#4 (``interactive navigation'') and item \#5 (``task and motion planning'')}

\rev{Generalising beyond the above task decomposition paradigm, where sub-tasks consist of only navigation or only manipulation actions, early work also considered settings where agents were required to dispatch multiple capabilities (e.g., manipulation, navigation, dialogue) within the same procedure or episode. This integrated setting, manifesting in such applications as manufacturing assistance and surgical robotics, represented a new challenge for planning-based approaches: agents quickly needed to contend with heterogeneous action spaces, longer task horizons, and combinatorial complexity in configuring the space of low-level actions. This setting was formalised as the problem of \textit{task and motion planning} (TAMP) \parencite{cambon2009hybrid, choi2009combining, wolfe2010combined, kaelbling2010hierarchical, kaelbling2013integrated, garrett2021integrated}, wherein algorithmic solutions define hierarchies across the different environment interaction types, with each type having its own unique execution primitive. These execution primitives were configured by way of a task decomposition, between a task-planning module (symbolic planning) and motion and/or manipulation planning modules (geometric planning). Many such solutions were plagued, still, with high search complexity, as early symbolic planners needed to enumerate all possible operations in a state, in order to expand the search tree. On the other hand, motion/manipulation planners were able to deal more effectively with geometry, but struggled when given partial goal specification. Notably, \textcite{kaelbling2010hierarchical} utilised \textit{goal regression}, to recursively decompose the planning problem, through aggressive hierarchicality---limiting the length of plans and, thereby, exponentially decreasing the amount of search required.\footnote{\rev{\textcite{garrett2021integrated} further formalise a class of \textit{task and motion planning} problems and survey solution methodology.}} With the advent of more sophisticated simulators (e.g., AI2-THOR \parencite{AI2THOR}) and benchmarks (e.g., ALFRED \parencite{ALFRED}, Interactive Gibson \parencite{iGibson20}), recent approaches extend the original TAMP problem in the context of \textit{mobile manipulation}, \textit{interactive navigation}, and \textit{navigation among movable objects}. \textcite{li2020hrl4in} pursue interactive navigation by proposing a hierarchical reinforcement learning framework, for learning cross-task hierarchical planning. \textcite{sharma2021skill} produced a framework for learning hierarchical policies from demonstration, with the intention of identifying reusable robot skills. Modularity as a general design principle in autonomous systems can be most closely attributed to the principles of cross-task hierarchical planning, as studied in this section, which serves as a strong foundation for improving the agents' sample-efficiency and the tractability of sometimes-conflicting learning objectives \parencite{das2018neural, chen2020learning}.}

\subsection{Learning Paradigms}

\subsubsection{Supervised Learning}\label{ssec:sl}

Supervision in \evlp~generally refers to a demonstration of a possible solution to a problem. Learning from demonstration approaches are typically focused on matching the behavior of their demonstrator or \textit{expert}  \parencite{hester2018deepql}. As such, these approaches can be suitable when there is available data from an expert, or when it is easier to collect expert demonstrations than to specify a reward function to train an agent under reinforcement learning. Furthermore, this type of learning paradigm has shown to help in accelerating the learning process in difficult exploration problems \parencite{syed2007gameil}.  

Different approaches within this paradigm have been used in all \evlp~tasks. For example, in VLN, \textcite{R2R} use the \textit{Professor-Forcing} \parencite{lamb2016profforc} approach, where at each training step the expert action is used to condition later predictions. Here, expert demonstrations generally correspond to the shortest-path from a start to a goal location for any given instruction. Furthermore, this approach is coupled with \textit{Student-Forcing}, a method which samples an agent's action from a output probability distribution in order to avoid limiting exploration to only the states included in the expert trajectories \parencite{R2R}. Another commonly used approach is DAgger \parencite{DAgger}, which trains on aggregated trajectories obtained using expert demonstrations \parencite{FollowingNavInstructionsQuadIL, VLN-CE, RoboVLN}. 

In EQA, \citet{eqa_matterport} train a behavior cloning model using a synthetic dataset of demonstration episodes, e.g., the shortest paths between an agent's location and the best viewpoint of an object of interest. In VDN, an approach known as Imitation Learning with Indirect Intervention (I3L)~\citep{nguyen2019i3l} has been used where, in addition to learning from expert demonstration during training, an agent can also request help from an \textit{assistant} both during training and evaluation in order to navigate to an object specified by the instructions \citep{nguyen2019hanna}.

In general, agents who learn from demonstrations do not generalize well, suffering from distribution shift issues due to the greedy nature of imitating expert demonstrations~\parencite{SQIL}. This tendency leads the agent in question to overfit to the seen environments, generally resulting in poor performance in the new, unseen environments.  

\subsubsection{Reinforcement Learning}\label{ssec:rl}

Although learning from supervision can lead to a faster learning process, agents trained under this paradigm often accumulate significant error due to limiting their exploration to the expert states \parencite{SERL}. In Reinforcement Learning (RL) settings, as opposed to having supervisions, an agent learns through interactions with an environment, e.g., by taking actions and receiving feedback from them. Through this setting, agents often learn more general behavior, and are capable of overcoming erroneous actions that may arise in unseen scenarios \parencite{LookLeap}.

\upd{Policy gradient methods are frequently used in embodied research. This type of algorithms directly models and optimises a policy function, which provides an agent with the guideline for the optimal action that it can take at a given state. Within embodied AI research, two popular types of policy gradient methods used include REINFORCE algorithms \citep{williams1992reinforce} and actor-critic algorithms \citep{mnih2016a3c, schulman2017ppo}. The former use episode samples to update an agent's policy parameters, whereas the latter combine policy learning with value learning \citep{sutton2018reinforcement}. Here, the policy has the role of the \textit{actor}, choosing the action to take, while the value function plays the role of the \textit{critic}, which criticizes the \textit{actor}'s decisions \citep{sutton2018reinforcement}.}

\upd{Actor-critic algorithms generally achieve smoother convergence and generally show superior performance, especially in continuous space problems \citep{schulman2017ppo}. One such example is Asynchronous Advantage Actor-Critic (A3C) \citep{mnih2016a3c}, an algorithm designed for parallel training, where multiple actors interact asynchronously within an environment being controlled by a global network.} \ann{Reviewer B, item \#6 (``PPO, DD-PPO'')} \rev{More recently, Decentralized Distributed Proximal Policy Optimization (DD-PPO) \parentcite{PointNav}, an algorithmic extension of PPO \citep{schulman2017ppo}, has gained popularity within embodied goal-oriented tasks such as PointGoal Navigation \citep{schulman2017ppo} and AudioGoal Navigation \citep{SoundSpaces}.}

\upd{Although the aforementioned techniques are used in \evlp~research, they have been used jointly with supervised learning methods which we discuss in the following section.} Nonetheless, some \evlp~approaches employ RL-based approaches exclusively. For instance, leveraging REINFORCE algorithms, \textcite{LookLeap} propose \textit{Reinforced Planning Ahead (RPA)} an approach that couples model-free and model-based RL to train a look-ahead model that allows an agent to predict a future state, and thus, plan before taking an action. 

For the tasks that require goal understanding and planning rather than instruction following, such as EQA and IQA, \citet{IQA} propose the \textit{Hierarchical Interactive Memory Network (HIMN)} where a model is decomposed into a hierarchy of controllers. For instance, a high-level planner chooses the low-level controllers where each low-level controller operates on a particular task, e.g., navigation, question answering, interacting, etc. Then, \citet{PredictiveQA} propose a Question Answering (QA) model that trains RL agents to use the notion of predictive modeling that stimulates the agents to visit places in the environment that might become relevant in the future. Although RL algorithms do not require labeled data and can endow agents with general skills to solve a task, they suffer from several difficulties including reward specifications, slow convergence, and sample-complexity.

\subsubsection{Joint Reinforcement and Supervised Learning}

Combining reinforcement and supervised learning is commonplace in \evlp~literature. 
By doing this, learning agents can leverage both the expert demonstrations and the direct feedback-based interaction with an environment to achieve more generalizable behavior. 

In VLN, \textcite{tan2019learning} use both imitation learning (IL) as a weak supervision to mimic expert trajectories generated using a shortest-path algorithm and on-policy RL to train a more general behavior. Then, \citet{wang2019reinforced} propose an RL model, \textit{Reinforced Cross-Modal Matching (RCM)}, where intrinsic rewards are used to encourage the learning agent to align the instructions with the trajectories. Then, they combine RCM with a self-supervised imitation learning approach (SIL), where the agent explores the environment by imitating its own past good decisions. \citet{NLMultitaskNav} use both RCM and behavioral cloning to develop a generalized multitask model for natural language grounded navigation tasks including VLN and VDN where learnable parameters are shared between the tasks. 

In the context of EQA, both \textcite{das2018neural} and \textcite{MultiAgentEQA} use IL to pre-train a model and RL to finetune it. \upd{In particular, \citet{das2018neural} propose \textit{Neural Modular Controller (NMC)}, which uses A3C to fine-tune a set of sub-policies trained through behavioral cloning to mimic expert trajectories from a global policy. In contrast, \textcite{MultiAgentEQA} uses REINFORCE to fine-tune an EQA agent trained on multi-target questions.} This strategy allows agents to learn to recover from errors, which  is not possible using expert demonstrations alone. 

\ann{Reviewer B, item \#6 (``PPO, DD-PPO'')}\rev{ \citet{blukis2019learning} introduce \textit{Supervised Reinforcement Asynchronous Leaning (SuReAL)} a framework that learns to map instruction to continuous control of quad-copters. SuReAL asynchronously trains two processes sharing data and parameters. The first one corresponds to a planner, trained via IL to predict a plan, and an action generator, trained using PPO to predict the control actions to solve a given plan. More recently, \citet{krantz2021waypoint} introduce waypoint models for VLN tasks in continuous environments. Motivated by the recent success of PointGoal tasks, \citet{krantz2021waypoint} leverage DD-PPO to extend the VLN-CE \citep{VLN-CE} task to support language-conditioned waypoint prediction models at varying action-space granularities.}

\citet{ActiveGathering} propose a module for exploration which enables an agent to decide when and what to explore. They first use IL to guide an initial exploration policy from an expert. Then, they use RL to explore the state-action space outside the demonstration paths and reduce the bias toward copying expert actions. To overcome the issue of reward specification in RL, \textcite{SERL} propose a VLN method based on Random Expert Distillation \parencite{RED}, known as Soft Expert Reward Learning (SERL), to learn a reward function by distilling knowledge from expert demonstrations. 

Another challenge that has been tackled by using RL and IL jointly is the accurate alignment of the instructions and visual features. \textcite{BabyWalk} propose the Babywalk approach which uses IL to first segment paths and then learn over these finer-grained, shorter instructions by demonstration. Babywalk then uses curriculum-based RL to refine the policy by giving the agent increasingly longer navigation tasks~\parencite{Curriculum}. This was shown to improve the model's ability to extrapolate to longer sequences, although generalization to unseen data does not appear to benefit as much from this approach.

\subsection{Common Techniques}\label{sec:common-tech}

The following subsections discuss common improvements beyond architecture, such as: data augmentation, pre-training, the use of additional training objectives, the use of different input representations, and optimization. Note that these can, and often are, used regardless of learning paradigm or architecture.

\subsubsection{Data Augmentation}
Data augmentation is commonly used to improve performance and robustness \parencite{ImageDataAugSurvey, tanner2012tools}. In \evlp, back-translation is the most commonly used data augmentation technique. Back-translation was originally introduced by the neural machine translation community \parencite{tan2019learning}. It augments an existing corpus by generating synthetic training samples for mono-lingual (i.e., single view) data. Due to the difficulty of creating a paired corpus, it is beneficial to train a ``backwards'' model which creates additional training pairs.

In VLN, paths are sampled throughout the environment. A speaker model is trained to generate synthetic instructions for a given path. These synthetic sentences supplement the training data \parencite{fried2018speaker}. Improvements on this idea were proposed in \cite{tan2019learning}, in which the authors introduced Environmental Dropout. Here, entire frames from ground-truth paths are omitted to create "synthetic" environments. New instructions are created for these synthetic environments using back-translation approach. This approach led to improvements in task success rate and narrowed the gap between the success rates on seen and unseen environments.

Other improvements to back-translation include changing the approach to sampling paths.\cite{ScenicRoute} proposes using longer paths which are not the shortest path between two points. Shortest paths may favor certain transitions in the training environment, which limit the agent's need to learn language. Instead, the paper proposes using a random walk which favors similar transition probabilities at each point in the graph. \citet{fu2020counterfactual} propose using counter-factual reasoning when sampling augmentation paths. An adversarial path sampler (APS) proposes increasingly hard paths for the model to train on. This module is model-independent and uses the training loss to choose paths the model is struggling with.

\subsubsection{Additional Objectives}
Additional tasks can also be used to increase the amount of training data and improve performance. In VLN, \citet{ProgressMonitor} add a completeness objective. Based on the data to date, the agent predicts how close to completion the agent is in meters. Progress monitor (PM) implicitly adds a prior over how instructions are processed. This approach encourages left-right attention of the instructions as the agent progresses through the path \parencite{ProgressMonitor}.

To further improve performance, \citet{AuxiliaryReasoningTasks} propose four additional tasks:
\begin{itemize}
    \item \textit{Trajectory retelling.} Generating a text based on the actions to date from the hidden state. The ground truth is generated using a speaker module \parencite{fried2018speaker}.
    \item \textit{Angle prediction.} Generate the angle in degrees of the agent's heading at the next time step.
    \item \textit{Instruction-path matching.} Shuffle states within a batch and see if they correspond to given instructions.
    \item \textit{Progress.} Percentage of steps taken.
\end{itemize}


Respectively, these auxiliary tasks have the following reasoning objectives: explaining the previous actions, predicting the next orientation, evaluating the trajectory consistency, and estimating the navigation progress \parencite{AuxiliaryReasoningTasks}.

\subsubsection{Pre-Training}
\label{subsec:pretrain}

Pre-training is done either by training a component of the model, or a full model on another problem entirely. This technique is used across most, if not all, neural \evlp~ approaches in the form of pre-trained image embeddings \parencite{Resnet}. Not only are image embeddings used for RBG images \parencite{CVDN, ALFRED,R2R,EQA,IQA}, but also for depth information \parencite{eqa_matterport, VLN-CE}.

Pre-training can also be used for language inputs. \cite{fried2018speaker} ablates the use of pre-trained word embeddings, finding that GloVe embeddings slightly improve performance \parencite{Glove}. More recent developments in language representation include pre-trained contextual models such as ELMO \parencite{Elmo}, BERT \parencite{devlin2018bert} or the GPT models \parencite{GPT2,GPT3}. \cite{li2019robust} encodes language using either BERT or GPT-2, giving significant improvements compared to a baseline learned LSTM encoding. 

Fusion of modalities can also benefit from pre-training. Recent approaches have taken advantage of large multimodal corpora \citep{ConceptualCaptions, MsCOCO, VisualGenome, shen2021clipbenefitvl} to pre-train BERT-like models \parencite{LXMERT, lu2019vilbert} which capture cross-modal information. \cite{VLNBERT} leverages pre-training to improve agent performance on VLN. It uses VilBERT \parencite{lu2019vilbert} as a base model to re-rank paths proposed by other models. \ann{Reviewer C, item \#3 (``Related Works'')} \rev{\citet{shen2021clipbenefitvl} integrate contrastive language-image pre-training (CLIP) \citep{radford2021clip}, a model designed for learning visual representations from natural language supervision, into Vision-and-Language tasks such as Visual Question Answering (VQA) \citep{VQA} and VLN. CLIP leverages a large amount of image-text pairs for training, achieving state-of-the-art performance with zero-shot transfer in a variety of computer vision tasks \citep{radford2021clip}. As such, \citet{shen2021clipbenefitvl} improve the performance of VLN agents on two datasets, R2R \citep{R2R} and RxR \citet{ku2020roomacrossroom} using EnvDrop \citep{tan2019learning} as their baseline model and CLIP as a visual encoder.}

\subsubsection{Multitask Learning}

A closely related topic to pre-training is multitask training. In the age of transformers this approach has grown in popularity for multimodal tasks \parentcite{lu202012, MTLHVR, MultitaskMultimodalEmotionAndSentiment}, including \evlp~tasks. 

\textcite{nguyen2019hanna} trains over three tasks datasets, two VLN datasets and a VDN dataset. The model proposed, \textit{PREVALENT}, is a transformer that fuses modalities in a later layer. It is pre-trained over speaker augmented R2R \parencite{fried2018speaker} by converting each step in the path to a \textit{action, image, text triple}. PREVALENT is trained to predict the action and masking over the instructions. The models are then specialized and fine-tuned over the R2R, CVDN, and HANNA datasets. This approach benefits CVDN in both seen and unseen environments, while improving performance uniquely over unseen environments for R2R and CVDN. \textcite{MultiTaskLearningVLN} train over both R2R and CVDN, but achieve lower performance over both datasets.

Task transfers have only been attempted between instruction following tasks. There have been no transfers between tasks requiring instruction following to those that require interpreting instructions. Doing so may be difficult, but could provide the a fine grained understanding of language, vision, and action required to interact with the world.

\subsubsection{Learning and Optimization}

Another way to improve performance is to modify the loss and optimization functions used. In \evlp~tasks, there are two ways to pose the loss function: as a sequence-to-sequence (Seq2Seq) problem; as a re-ranking problem. In a Seq2Seq problem, a series of inputs are used to predict a sequence of ground-truth actions. Current approaches treat this as a classification problem, with the goal of learning a maximum likelihood estimate using cross-entropy loss conditioned on the agent's observations. In a re-ranking problem, a Seq2Seq model first proposes candidate solutions, and a second model scores them. \citet{VLNBERT} does this through the use of negative sampling, creating binary labels which reflect the validity of a path. Binary-cross entropy is again used as an objective.

Reasons to choose one over the other include performance, computational cost, and whether or not the environment is known. Re-ranking problems are, at the time of writing, state of the art in VLN \parencite{VLNBERT}. However they are more computationally expensive and unable to explore the environment without use of an additional model(s).

Two optimization algorithms are commonly used \parencite{MultiModalSurvey}: ADAM \parencite{adam}, RMSPROP \parencite{rmsprop}. However the advantages or reasoning behind choosing one or the other are not discussed in \evlp~literature.
\ann{Reviewer B, item \#7 (``Reward Shaping'')} \rev{\subsubsection{Reward Shaping}}

\rev{Reward-shaping is a widely-used technique in solving Markov decision processes, as with Reinforcement Learning (RL), where domain knowledge is used to define a shaped reward $R_{shaped}$, for improved agent learning and optimisation \citep{RewardShaping}. Typically, this shaped reward is linearly combined with the original reward $R$, with $R^{'} = R + R_{shaped}$, or used instead. Reasons for incorporating a shaped reward include situations where $R$ is sparse or is deemed not sufficiently informative for the task or for encouraging the desired agent behaviour \citep{yun2018rewardshaping}.}

\rev{The \evlp~literature generally follows this standard formulation and adopts similar usage considerations for shaped rewards. In EQA, \citet{das2018neural}, \citet{IQA} and \citet{MultiAgentEQA} use a reward function $R$ of the form $R = R_{terminal} + R_{shaped}$, where $R_{terminal}$ is a sparse reward assigned for correctly answering a question. \citet{IQA} additionally assign a penalty for incorrect answers. $R_{shaped}$ consists of a dense reward determined by the agent's progress toward its goal: positive if moving closer to the goal and negative otherwise.}

\rev{VLN and VDN works \citep{LookLeap, tan2019learning, MultiTaskLearningVLN, ActiveGathering} have used a similar reward function where $R_{terminal}$ is given if the agent stops within a specified radius of the target, and $R_{shaped}$ is given as progress-based reward. \citet{wang2019reinforced} use the aforementioned rewards to define $R_{extr}$, an extrinsic reward to measure the agent's success ($R_{terminal}$) and progress ($R_{shaped}$), and further define $R_{intr}$, an intrinsic reward for measuring the alignment between the instruction and the agent's trajectory. The latter is obtained as a probability score of generating the original instruction, given sequential encodings of the instruction and the agent's historical trajectory. The total reward function was defined as $R = R_{extr} + \delta R_{intr}$, where $\delta$ is a weighting parameter.}

\rev{In addition to the sparse extrinsic reward, \citet{R4R} and \citet{BabyWalk} use fidelity of the predicted path with the reference path based on the Coverage weighted by Length Score (CLS) metric \citep{R4R}, discussed in Section \ref{ssec:path2path}, to shape the reward function. Similarly, SERL \citep{SERL} uses two intrinsic rewards, $R_{SED}$, a soft-expert distillation reward, and $R_{SP}$, a self-perceiving reward: the former is learned by aligning the VLN agent's behavior with expert demonstrations \citep{RED}, and the latter is learned by predicting the agent's progress toward its goal.}

\section{Evaluation} 
\label{sec:evaluation}

Environments and metrics form the backbone of \evlp~agent assessment. Proper evaluation systems allow to quantify specific aspects of agent performance (e.g., the ability to navigate toward a goal). Standardization of such metrics and the environments used to train and evaluate agents enables researchers to better coordinate current and future work \citep{SPL}. This section serves to summarize and compare the evaluation metrics and environments that are most relevant to the \evlp~tasks defined in Section \ref{sec:tasks}.

\subsection{Metrics}

The metrics currently used in \evlp~can be grouped into five categories, each measuring a different aspect of agent performance: (i) success, (ii) distance, (iii) path-path similarity, (iv) instruction-based metrics, and (v) object selection metrics. We survey and discuss the existing methods for each category. For additional ease of reference, we summarize these metrics in tables \ref{tab:metrics-success}-\ref{tab:metrics-pif}. 

\subsubsection{Success}

Metrics that assess whether or not an agent successfully completed a task are used, in various forms, across all \evlp~tasks. Instruction-following tasks such as VLN and VDN typically define the \textit{success rate (SR)} as how often  the agent gets within a threshold distance $d_{th}$ of the goal. Distance $d_{th}$ varies between datasets, with $d_{th} = 3m$ for R2R \parencite{Matterport3D}, and $d_{th} = 0.47m$ for Lani \parencite{misra2018mapping}. This measure has some weaknesses, particularly for discretized state-action spaces, notably that it is rather dependent on the granularity of said discretization \parencite{UnimodalBaselines, VLN-CE}. Furthermore, variations in $d_{th}$ also impact SR, which may prompt misleading results where correct executions might be considered wrong \parencite{blukis2019learning}.

EQA has no notion of a target location; instead, success is measured in terms of accuracy over the output space, i.e., question-answering accuracy \parencite{EQA,IQA}. Other common classification metrics that can be used include precision, recall, and \textit{F-score} \parencite{Valuations2015ARO}. Per-class performance for any of these metrics can also be employed to measure if the model is improving accuracy by outputting common responses at the cost of performance over minority classes.

In most cases, a task is considered as ``complete" only after the agent is near a final destination, answers a question, or satisfies a manipulation directive. However, certain datasets such as ALFRED \parencite{ALFRED}, Lani \parencite{misra2018mapping}, and CerealBar \parencite{CerealBar} report sub-goal completion as well. This addition of sub-goal scoring can be useful, for instance, to exploit modularity and hierarchy of sub-tasks \citep{ALFRED, jansen-2020-visually}, or to alleviate issues with success metrics, e.g. that early failures early in a task impact later steps in the episodes \citep{CerealBar}, making it difficult to assess the ability of an agent per timestep. To address this issue, CerealBar evaluates a trajectory $T$ as $|T| - 1$ sub-paths, each shorter than the next: the average success rate across these runs is taken, measuring the model's ability to recover from errors. A further benefit of sub-goal scoring is that specific agent mistakes can be pinpointed and categorized more easily.

\subsubsection{Distance}

While success is an intuitive measure, it provides limited information on how efficiently the agent traveled through a space. Distance measures can quantify this type of information, in the context of both navigation and manipulation tasks. Two useful metrics in this respect are navigation (or displacement) error (NE) and overall amount of travel (or motion) required known as the path length (PL). NE measures how far from a target location an agent chooses to stop, giving information about the magnitude of the agent's inability to get to the target location. In order to further analyze whether agents learn to stop effectively, NE is often coupled with an Oracle Navigation Error (ONE), which measures the distance of the nearest point along the agent's trajectory to the goal location. Significantly different NE and ONE measures may indicate an issue with stopping.

These measures assume that the task has a specified goal or target location. In the context of VLN and VDN this is a sensible assumption. However for EQA, or even EGM, that is not necessarily the case. For example, in EQA, certain questions such as those about the state of an object, the answer could be given from different points in the room. Rather than assess the distance from a goal, one could look at the length of the path taken commonly referred to as PL. In a real-world scenario, minimizing the length of travel is an important consideration. To only account for successful paths, SR weighted by normalized inverse Path Length (SPL) can be calculated.

In essence, distance can be used not only to derive success metrics for VLN, VDN, but also to quantify the error and efficiency of the agent's path. In the context of navigation, this measures how much does the agent meander from its target. This adds a dimension when evaluating and comparing model performance.

\subsubsection{Path-Path Similarity}
\label{ssec:path2path}
Distance metrics fail to capture whether or not the agent adhered to the ground-truth path. For instruction-following tasks such as VLN and VDN, where the user specifies a given path with instructions, we may want to explicitly follow those directions and not find the shortest path. \textcite{R4R} discuss desiderata for path similarity metrics:

{\it
\begin{enumerate}
    \item Metrics should gauge the fidelity between the path taken and the reference path, rather than only the goal
    \item Mistakes should not be hard penalties, metrics should favor a path that deviates and follows the ground truth path over one that short cuts it entirely
    \item Maxima should be unique and only occur in the case of a perfect match
    \item The underlying scale of the dataset should not impact the value of the output
    \item Metrics should allow fast automated performance evaluation
\end{enumerate}
}
 
 We believe these desiderata can be applied more widely to any instruction-following task. One of the early metrics to measure path following is edit distance \parencite{TouchDown}. Edit Distance (ED) measures the number of changes required to match two graphs. ED does not satisfy criterion \#2, as it only penalizes based on absolute deviation. \textcite{R4R} introduce Coverage weighted by Length Score (CLS), a metric which satisfies all 5 of these criteria. CLS is given by the product of the path coverage, a measure of average per point overlap between the target and reference path, and the length score, which penalizes paths both shorter and longer than the reference. \citet{R4R} verify its utility by using it as an objective for a RL-based agent.
 
 \textcite{nDTW} build on this work and propose two measures: normalized Dynamic Time Warping (nDTW) and Success weighted by normalized Dynamic Time Warping (SDTW). The former is a similarity function that identifies an optimal warping of the elements from a reference path and a query path, while the latter constrains nDTW to successful episodes. These metrics have a number of desired properties: they respect the desiderata outlined above, they can be used effectively as a reward signal for RL-based agents, and can be compared with human judgment. Such properties make them consistently preferred compared to proposed alternatives such as CLS, SPL, SR.

\subsubsection{Instruction-Based}

Different approaches can be used to measure the alignment between paths in the form of images and text: Semantic Propositional Image Caption Evaluation (SPICE) \parencite{anderson2016spice} and Consensus-based Image Description Evaluation (CIDEr) \parencite{vedantam2015cider}. While those were originally used in the context of VQA, they have been recently adapted to VLN as well \parencite{OnEvalVLN}. 

An alignment metric can also be used as part of training. Certain approaches \parencite{fried2018speaker, tan2019learning} generate synthetic instructions to augment the data available, of varying quality. \textcite{Discriminator} train an alignment model in which instructions and paths are encoded using BiLSTMs and fed into a binary classifier. Positive outputs are assigned to matched path instruction pairs and negative outputs are mismatched instruction path pairs. This binary classifier is then used as a scoring function to rank paths. Low scoring paths are removed from training leading to a statistically significant improvement in performance while reducing the total dataset size.

\subsubsection{Object Referral}

EGM and EOR require object selection, either through masking or selecting from ground-truth bounding boxes. \textcite{TouchDown} and \textcite{ALFRED} use Intersection over Union (IoU) as an evaluation metric, a commonly-used measure in CV, specifically in object detection tasks \parencite{ObjectDetectionMetricSurvey}. However, it is not the only metric nor the most informative one. 

\begin{table}[!htp]
\rowcolors{3}{white}{gray!10}
\caption{\footnotesize List of \textit{success} metrics used in \evlp~ tasks with a brief description, type of metric, types and references to tasks they are typically used in, and the formula if applicable. The nomenclature used is as follows: $P$ is the predicted path, $R$ is the reference path corresponding to $P$, $p$ and $r$ are nodes in $P$ and $R$, respectively. $d$ is the appropriate distance, $S$ is the set of instruction-path pairs.  $d_{th}$ is threshold distance, $BB$ is the bounding box, $\hat{y}$ is a prediction value, and $y$ the corresponding ground-truth value.}
\label{tab:metrics-success}
\scriptsize
\resizebox{1\columnwidth}{!}{
\begin{tabularx}{1.052\textwidth}{
p{0.26\textwidth}
p{0.19\textwidth}
p{0.06\textwidth}
p{0.19\textwidth}
p{0.215\textwidth}
}
\toprule
\textbf{Metric} & \textbf{Description} & \textbf{Type} & \textbf{Typical Tasks} & \textbf{Formulas} \\
\midrule
SR \citep{R2R} & 
Percentage of episodes an agent stops within a threshold distance of goal &
Success & 
VLN, VDN, EOR, EGM & 
$ \frac{1}{|S|}\sum\limits_{P, R \in S} \mathds{1}[NE(P,R)< d_{th}]$ \\
OSR \citep{R2R} & 
SR if the agent had stopped at it's nearest point to the goal & 
Success & 
VLN, VDN, EOR & 
$ \frac{1}{|S|}\sum\limits_{P, R \in S} \mathds{1}[ONE(P,R)< d_{th}]$ \\
Sub-goal SR \citep{ALFRED} & 
SR for a specific sub-goal & 
Success & 
VLN, EGM & 
$SR$ \\
Accuracy \citep{EQA} & 
Measures whether a prediction value matches the ground-truth value &
Success & 
EQA & 
$\mathds{1}(\hat{y} = y)$ \\
EM & 
The agent perfectly replicates the target path & 
Success & 
All & 
$\mathds{1}(P=R)$ \\ 
\bottomrule
\end{tabularx}
}
\end{table}
\begin{table}[!htp]
\rowcolors{3}{white}{gray!10}
\caption{\footnotesize List of \textit{distance} metrics used in \evlp~ tasks with a brief description, type of metric, types and references to tasks they are typically used in, and the formula if applicable, where $P$ is the predicted path, $R$ is the reference path corresponding to $P$, $p$ and $r$ are nodes in $P$ and $R$, respectively.}
\label{tab:metrics-distance}
\scriptsize
\resizebox{1\columnwidth}{!}{
\begin{tabularx}{1.108\textwidth}{
p{0.22\textwidth}
p{0.21\textwidth}
p{0.07\textwidth}
p{0.19\textwidth}
p{0.28\textwidth}
}
\toprule
\textbf{Metric} & \textbf{Description} & \textbf{Type} & \textbf{Typical Tasks} & \textbf{Formulas} \\
\midrule
NE \citep{misra2018mapping} & 
Geodesic distance between an agent's final position and the goal & 
Distance & 
VLN, VDN & 
$d(p_{|P|}, r_{|R|})$ \\
ONE \citep{misra2018mapping} & 
Distance to target if the agent has stopped at the nearest point & 
Distance & 
VLN, VDN & 
$\min_{p\in P} d(p,r_{|R|})$  \\
PL \citep{R2R} & 
How far the agent traveled & 
Distance & 
VLN, VDN, EOR &  
$ \sum\limits_{i=1}^{|P|-1} d(p_i, p_{i+1}) $ \\
SPL \citep{SPL} & 
Average of the path lengths where the agent succeeded & 
Distance & 
VLN, VDN, EOR, EGM & 
$ \mathds{1}[NE(P,R)< d_{th}]\frac{d(p_1, r_{|R|})}{\max (PL(P), d(p_1, r_{|R|}))} $ \\
\bottomrule
\end{tabularx}
}
\end{table}
\begin{table}[!htp]
\rowcolors{3}{white}{gray!10}
\caption{\footnotesize List of \textit{path-path similarity} metrics used in \evlp~ tasks with a brief description, type of metric, types and references to tasks they are typically used in, and the formula if applicable; here, $P$ is the predicted path, $R$ is the reference path corresponding to $P$, $p$ and $r$ are nodes in $P$ and $R$, respectively. $d$ is the appropriate distance, $d_{th}$ is threshold distance, and $lev(\cdot, \cdot)$ is the Levenshtein edit distance.}
\label{tab:metrics-pps}
\scriptsize
\resizebox{1\columnwidth}{!}{
\begin{tabularx}{1.12\textwidth}{
p{0.21\textwidth}
p{0.25\textwidth}
p{0.12\textwidth}
p{0.12\textwidth}
p{0.28\textwidth}
}
\toprule
\textbf{Metric} & \textbf{Description} & \textbf{Type} & \textbf{Typical Tasks} & \textbf{Formulas} \\
\midrule
CLS \citep{R4R} & 
Measure of similarity between paths using path coverage and length-score &
Path-Path Similarity & 
VLN & 
$PC(P,R) \cdot LS(P,R);$ \par $PC(P,R) = \frac{1}{|R|} \sum\limits_{r\in R} \exp{\Big(\frac{-d(r,P)}{d_{th}}\Big)};$ \par $LS(P,R) =\frac{PC(P,R) \cdot PL(R)}{PC(P,R) \cdot PL(R) + |PC(P,R) \cdot PL(R) - PL(P)|}$ \\
nDTW \citep{nDTW} & 
Dynamic time warping normalized between target and reference path & 
Path-Path Similarity & 
VLN & 
$\exp{\Big(-\frac{DTW(R,P)}{|R| \cdot d_{th}}\Big)}$ \\
SDTW \citep{nDTW} & 
nDTW weighted by success rate & Path-Path Similarity & 
VLN & 
$\mathds{1}[NE(P,R)< d_{th}] \cdot nDTW(P, R)$ \\
SED \citep{TouchDown} & 
Success weighted edit distance & Path-Path Similarity & 
VLN & 
$ \mathds{1}[NE(P,R)< d_{th}] \cdot ( 1 - \frac{lev(P,R)}{\max(|P|,|R|)})$ \\
\bottomrule
\end{tabularx}
}
\end{table}
\begin{table}[!htp]
\rowcolors{3}{white}{gray!10}
\caption{\footnotesize List of \textit{path-instruction fidelity} metrics used in \evlp~ tasks with a brief description, type of metric, types and references to tasks they are typically used in, and the formula if applicable. The nomenclature used is as follows: $S$ is the set of instruction-path pairs, $BB$ is the bounding box, $\hat{y}$ is a prediction value, and $y$ the corresponding ground-truth value. Entries with a single asterisk (*) are learned approaches. For entries with double-asterisks (**), we refer the reader to the referenced work.}
\label{tab:metrics-pif}
\scriptsize
\resizebox{1\columnwidth}{!}{
\begin{tabularx}{1.174\textwidth}{
p{0.255\textwidth}
p{0.21\textwidth}
p{0.15\textwidth}
p{0.14\textwidth}
p{0.28\textwidth}
}
\toprule
\textbf{Metric} & \textbf{Description} & \textbf{Type} & \textbf{Typical Tasks} & \textbf{Formulas} \\
\midrule
BLEU \citep{BLEU} & 
Compare generation caption to some reference & Path-Instr. Fidelity &
Instr.~generation &
$BP\cdot\exp{\big(\frac{1}{|N|}\sum\limits_{n=1}^{N} \frac{\sum count(n,s)}{\sum count(n,\hat{s})}\big)}, s \in S;$ \par $BP = \exp{\big(1 - \frac{|s|}{|\hat{s}|}\big)} \texttt{ if } |\hat{s}| \leq |s| \texttt{ else } 1$ \\
SPICE \citep{anderson2016spice} & 
Measure of image-caption alignment & 
Path-Instr. Fidelity & 
Instr.~generation & 
* \\
CIDER \citep{vedantam2015cider} & Parsing based method to align captions to images & Path-Instr. Fidelity & Instr.~generation & ** \\
ROUGE-N \citep{ROUGE} & 
Compare caption generation to some reference & 
Path-Instr. Fidelity & Instr.~generation & 
$\frac{ \sum_{n-gram \in s} Count_{match}(n-gram)}{\sum_{\hat{n-gram} \in \hat{s}} Count(\hat{n-gram})},s \in S$ \\
Path-Instr \par discriminator \parencite{Discriminator} & 
Discriminator trained to find matching vs synthetically sampled datasets &
Path-Instr. Fidelity & 
VLN & 
** \\
Path-Instr \par compatibility \citep{OnEvalVLN} & Compares path-instruction alignment & Path-Instr. Fidelity & VLN & * \\
IoU \citep{VOC} & Measures overlap between 2 bounding boxes & Object Referral & EQA, EGM, EOR &  $ \frac{|BB_1 \cap BB_2 |}{| BB_1 \cup BB_2|}$ \\
\bottomrule
\end{tabularx}
}
\end{table}

Recently, \textcite{ArraMon} proposed three metrics to evaluate how well objects are selected: Collected Target Correctness (CTC), Placement Target Correctness (PTC), and reciprocal Placement Object Distance (rPOD). CTC measures if the correct object was selected; PTC measures if the object was correctly placed. Note that CTC and PTC can both be framed as thresholded IoU problems, as that is how some agents \parencite{ALFRED} decide whether or not to pick up an object. rPOD is a normalized measure that gives a near zero score if the object is far from it's ideal placement and a score of 1 if the placement is perfect.

While only applicable to EGM and EOR, measuring the quality of object selection interaction is important. It is another aspect of the task and provides a good way of understanding what aspects of language are harder for the model. Recent work in object recognition \parencite{ObjectDetectionMetricSurvey} reveals that there are many metrics and variants. Even more traditional metrics such as precision, recall, and F-score are not currently reported in those tasks. Doing so would provide more information about the bounding box accuracy. In addition we could report per-class accuracy, which would be useful to identify which objects are poorly detected by the model. 

\subsection{Simulation Environments and Datasets} 
\label{ssec:datasets}

Solving \evlp~tasks generally involves using a simulation environment and a dataset. Simulation platforms and datasets facilitate the reproduction and evaluation of embodied systems. Simulators aim to replicate aspects of the real-world and model agents capable of solving complex tasks while abstracting away the difficulties of designing and supervising a real-world agent. In contrast, datasets play a crucial role in articulating how each task is framed. They provide examples of \textit{how} agents should behave in response to certain multimodal stimuli.

In this section, we provide a per-task summary and comparison of literature pertaining to environments and datasets used in, or designed for \evlp~research. Furthermore, we provide a brief discussion of other recent and relevant multimodal platforms and datasets outside the \evlp~tasks discussed in this survey article. Finally, we summarize important aspects of these simulators and datasets in Table~\ref{Tab:SimTask} and Table~\ref{tbl:dataset}, respectively.

\subsubsection{Simulators}
\label{ssec:simulators}
\noindent Early simulation platforms for embodied research typically leverage video game environments to create and train neural controllers. Human performance was quickly achieved on some of these platforms \citep{ATARIEnv, ViZDoom, TorchCraft, DeepMindLab, ProjectMalmo, Retro} as simplified environments generally lack the diversity and complexity of real-world settings \parencite{MINOS, AI2THOR, RoboTHOR}. 

Recent works have addressed this lack of realism through the use of photo-realism \citep{AIHabitat, GibsonEnv, R2R, herman2021learn} and the use of interactive contexts where agents are able to modify the states of objects in the environment \citep{AI2THOR, iGibson, CHALET, HoME}. Toward this end, there is also interest in developing frameworks focused on simulation-to-real transfer and evaluation \citep{RoboTHOR}, allowing the study of discrepancies between real settings and simulated ones. Finally, other platforms have also focused on encouraging reproducibility of work, flexibility of design, and benchmarking \citep{AIHabitat, AllenAct}.

\para{VLN Simulators.} Three environments are relevant to VLN tasks: Matterport3DSim \parencite{Matterport3D, R2R}, Habitat \parencite{AIHabitat}, and StreetLearn \parencite{mirowski2019learning}. In both Matterport3DSim and Habitat, VLN is set in indoor environments, while StreetLearn is set outdoors. Matterport3DSim allows an agent to navigate in a discrete environment framed as an undirected navigation graph which associates positions with the corresponding viewpoints, subject to admissible (i.e., collision-free) intra-node transitions. Agents navigating in this setting \textit{teleport} between viewpoints, through high-level atomic actions.

In contrast, \textcite{VLN-CE, RoboVLN} leverage Habitat to provide a more realistic setting for VLN, where agents can navigate in a a discrete \parencite{VLN-CE}, as well as, continuous \parencite{RoboVLN} unconstrained state-space, as opposed to a discretized graph. In the unconstrained setting, agents execute low-level actions over longer horizon sequences \parencite{RoboVLN}, and more realistically, risk colliding with obstacles along the way. Tangentially, VLN has also been studied in the context of simulation-to-real transfer and evaluation. \textcite{Sim2RealVLN} ports a VLN agent trained in Matterport3DSim using Robot Operating System (ROS) as a publish-subscribe framework for robot control. They also evaluated the gap in performance between an agent moving through high-level actions compared to a robot taking low-level actions in the real world. They found that, although there has been significant progress in simulation environments for VLN, simulation-to-real transfer is still not reliable, particularly when the agent does not have access to a map.

\para{EQA Simulators.} Both, indoor, synthetic \parencite{EQA, IQA} and photo-realistic \parencite{eqa_matterport} environments are used for EQA. \textcite{EQA} use House3D \parencite{wu2018building}, a synthetic large-scale environment to train EQA agents, whereas \textcite{IQA} used AI2-THOR \parencite{AI2THOR}. In both settings, the agent navigates in a discrete unconstrained environment and perceives the world through ego-centric RGB images. While AI2-THOR is not a large-scale environment as House3D, it enables the study of interactive tasks something crucial in IQA.  

\textcite{eqa_matterport} argue that although the above-mentioned environments are semantically realistic in terms of layout and types of surrounding objects, they lack visual realism in terms of graphics and intra-class variation. Thus, they propose MP3D-EQA, an EQA task set in a photorealistic simulator, and built upon the MINOS simulator \parencite{MINOS} and the Matterport3D dataset \parencite{Matterport3D}. In addition leveraging RGB images, they add 3D point-cloud observations, which they empirically showed to be more effective for navigation, in contrast to previous works. More recently, MP3D-EQA has also been developed in Habitat. \textcite{AIHabitat} showed that Habitat's performance in terms of frames-per-second (FPS) is significantly superior to that of MINOS, and other platforms, allowing to shift bottlenecks in simulation training. 

\para{EOR Simulators.} Both indoor and outdoor settings exist for the EOR task. REVERIE \parencite{qi2020reverie} is an indoor-based simulator built upon Matterport3DSim. In particular, it adds object-level bounding boxes to Matterport3DSim allowing for object retrieval. Since a target object may be observed from multiple viewpoints, and because it is expected that the agent reaches the goal within a short distance, they only preserve object bounding boxes within a 3-meter radius of any given viewpoint. 

TouchDown \parencite{TouchDown} was made publicly available by transferring it to the StreetLearn  environment \parencite{mirowski2019learning,mehta2020retouchdown}. StreetLearn contains panoramas from New York City and Pittsburgh. Like Matterport3DSim, navigation in StreetLearn is based on an undirected graph, and agents move through discrete actions.

\para{VDN Simulators.} The VDN task family has also been established in Matterport3DSim \parencite{nguyen2019hanna, CVDN, JustAsk}. Typically the VLN task is extended with crowd-sourced dialog or sub-instructions that provide the agents with assistance to complete the task. We refer the reader for further information to Section~\ref{sssec:datasets}.

\para{EGM Simulators.} As discussed in Section \ref{subsubsec:egm}, interaction with the environment is a crucial component of the EGM task family. Currently, photo-realistic simulation environments are limited in providing interactive or dynamic interaction prospects. Consequently, EGM tasks have mainly leveraged synthetic environments \parencite{ALFRED, misra2018mapping, CerealBar, ArraMon}. 

\textcite{ALFRED} used the AI2-THOR 2.0 simulation environment for the ALFRED task. In contrast to IQA, which also leverages this simulator, ALFRED does not require agents to answer contextualized questions.  Instead, agents must complete tasks that require changing the state of objects (e.g., cleaning, picking, stacking, and heating objects). 

In contrast, ArraMon \parencite{ArraMon}, and CerealBar \parencite{CerealBar} are both framed in outdoor synthetic environments. ArraMon uses a custom two-phase environment consisting of synthetic street-layouts and buildings, used for the navigation phase, and a grid-based room layout used during the assembly phase. CerealBar was designed to study of collaborative interactions between agents. 
\ann{Reviewer B, item \#3 (``Gibson'')} \ann{Reviewer C, ``Related Works", item \#2}\rev{\paragraph{Other Simulators.} We briefly discuss other platforms that have recently shown significant promise, in the definition of multimodal embodied tasks that outside conventional \evlp~families. Some of the features proposed by these simulators include designing interactive tasks with improved physics \citep{iGibson, szot2021habitat20, SAPIEN, TDW} and object states \citep{iGibson20}, additional modalities \citep{SoundSpaces, iGibson10}, low-level articulation \citep{SAPIEN, szot2021habitat20}, and integraton with virtual reality interfaces \citep{iGibson20, TDW}}.

\rev{iGibson 1.0 \citep{iGibson, iGibson10} is a recent simulation environment that allows for the development of interactive navigation and manipulation tasks in large-scale realistic scenes, in contrast to its precursor Gibson \citep{GibsonEnv}, which only provides support for robot navigation tasks (where agent interaction is limited to collisions with scene meshes). Among the most valuable aspects of this platform are the physics-based simulation capabilities, which enable high-quality interactive tasks. More recently, iGibson 2.0 \citep{iGibson20} was provided as an extension to iGibson 1.0; it includes a set of extended states and logical descriptions, which  simulate such underlying physical processes as temperature, wetness, and cleanness of objects---thus motivating a wider range of household tasks and perceptual modes. Moreover, to facilitate modeling household tasks, iGibson 2.0 introduces a virtual reality (VR) interface to allow humans to collect demonstrations, e.g., for developing solutions that leverage imitation-based learning. } 

\rev{\citet{szot2021habitat20} present Habitat 2.0, a simulation platform for training virtual robots in interactive environments and physics-enabled synthetic scenarios. It was built upon Habitat \citep{AIHabitat} and, as such, it prioritises high-performance and speed. \citet{szot2021habitat20} introduce the Home Assistant Benchmark (HAB) a suite of tasks for assistive robots, such as, preparing groceries and cleaning the house, designed for testing the manipulation capabilities of agents.} \upd{ Also extending Habitat, SoundSpaces \citep{SoundSpaces} integrates environment acoustics to enable audio-based navigation tasks in indoor environments. To do so, they leverage algorithms for modeling room acoustics as well as sound reflections based on room geometry.}

\upd{VirtualHome \citep{VirtualHome} is an environment designed for simulating household activities by generating programs containing sequences of symbolic instructions representing atomic actions. Modalities supported by this platform include RGB, depth, semantic segmentation and natural language.}

\upd{ThreeDWorld (TDW) \citep{TDW} enables the development of interactive navigation and manipulation tasks in both outdoor and indoor environments. It also allows users to procedurally create custom environments (and the objects to populate them), as well as interact with objects through VR. TDW also supports complex physical dynamics, with audio support, which is relevant for exploring topics such as physical reasoning \citep{SMaterialSound} and audio-visual learning \citep{SoundSpaces, Ssstereo}. Finally, Sapien \citep{SAPIEN} is a synthetic-based platform that provides rich physics support for developing low-level interactive manipulation tasks.}

\begin{table}[ht]
\rowcolors{2}{gray!10}{white}
\caption{\ann{Reviewer B, item \#3 (``Gibson'')} \ann{Reviewer C, ``Related Works", item \#2}\ann{Reviewer C, ``Minor Comments", item \#1} \rev{A comparison of simulation environments, related to \evlp~research. \textbf{Supported Tasks:} we include PointNav \citep{PointNav}, RoomNav\citep{wu2018building}, AVN (Audio-Visual Navigation; \citep{SoundSpaces}), HAB \citep{szot2021habitat20}, ArmPointNav \citep{ehsani2021manipulathor}, general navigation, and manipulation, in addition to conventional \evlp~tasks. \textbf{Agent-World Interaction:} C: continuous actions / forces, D: discrete actions. \textbf{Environments:} S: synthetic, P: photorealistic. \textbf{Scenes:} I: indoors, O: outdoors. \textbf{Modalities:} A: audio, C: color, D: depth, F: flow, L: LiDAR, NL: natural language, SN: surface normals, SS: semantic segmentations. \textbf{Scene Scale:} B: buildings, R: rooms, H: houses, RO: realistic outdoors, SO: synthetic outdoors O: object-level. }}
\resizebox{\columnwidth}{!}{
\centering
\begin{tabular}{l c c c c c c c c c c} 
    \toprule
    \textbf{Environment} & 
    \upd{\textbf{Supported Task}} &
    \upd{\begin{tabular}[c]{@{}c@{}} \textbf{Agent-World}\\ \textbf{Interaction}\end{tabular}} &
    \textbf{Environments} &
    \textbf{Scenes} &
    \textbf{\rev{Scene Scale}} &
    \textbf{\upd{Modalities}} &
    \textbf{Multi-Agent} &
    \upd{\textbf{Specialty}} &
    \textbf{Reference} \\
 \midrule
 Matterport3DSim 
 & VLN, VDHN, EOR      
 & D 
 & P 
 & I 
 & B 
 & C,D,NL 
 &
 & Topological planning
 & \citet{Matterport3D}  \\
 MINOS 
 & EQA, PointNav, RoomNav       
 & C,D 
 & P,S 
 & I 
 & B 
 & C,D,NL,SN,SS 
 & 
 & High-speed, navigation
 & \citet{MINOS}\\
 AI2-THOR  
 & VDHN, EQA, EGM 
 & C,D 
 & S 
 & I 
 & R 
 & A,C,D,SS 
 & \cmark 
 & Actionable objects, task planning
 & \citet{AI2THOR}\\
 ManipulaTHOR  
 & ArmPointNav 
 & D 
 & S 
 & I 
 & R 
 & C,D
 &  
 & Mobile manipulation
 & \citet{ehsani2021manipulathor}\\
 House3D
 & EQA, RoomNav            
 & D 
 & S 
 & I 
 & B,R 
 & C,D,SS,NL 
 & 
 & Environment diversity, navigation
 & \citet{wu2018building}\\
 CerealBar 
 & EGM            
 & D 
 & S 
 & O  
 & SO 
 & C,NL 
 & \cmark
 & Collaborative interactions
 & \citet{CerealBar} \\
 ArraMon 
 & EGM            
 & D 
 & S 
 & O 
 & SO 
 & C,NL 
 & 
 & Joint navigation and manipulation
 & \citet{ArraMon} \\
 StreetLearn      
 & VLN, EOR            
 & D 
 & P 
 & O 
 & RO 
 & C,NL 
 & 
 & Urban Navigation
 & \citet{mirowski2019learning}\\
 Gibson 
& Nav.
& C 
& P,S 
& I 
& H 
& C,D,SN,SS 
& 
& Navigation
& \citet{GibsonEnv} \\
\rev{iGibson 1.0}
& Nav., Manip.
& C 
& P,S 
& I 
& H 
& C,D,F,L,SN,SS 
&  
& Physics Interaction
& \citet{iGibson10} \\
\rev{iGibson 2.0}
& Nav., Manip.
& C 
& P,S 
& I 
& H 
& C,D,F,L,SN,SS 
&  
& Extended object states, VR
& \citet{iGibson20} \\
 Habitat 
 & VLN, EQA, PointNav      
 & C,D 
 & P,S 
 & I 
 & B 
 & C,D,SS 
 & 
 & High-speed, navigation, customizable
 & \citet{AIHabitat} \\
 \rev{Habitat 2.0}
 & HAB      
 & C 
 & S 
 & I 
 & H 
 & C,D 
 & 
 & High-speed, task planning, articulated dynamics
 & \citet{szot2021habitat20} \\
 \upd{SoundSpaces}
 & AVN      
 & C,D 
 & P,S 
 & I 
 & B 
 & A,C,D,SS 
 & 
 & High-speed, audio-visual navigation
 & \citet{SoundSpaces} \\
\upd{VirtualHome}
& Nav., Manip.
& D 
& S 
& I 
& H,R 
& C,D,F,SS 
& 
& Task Planning
& \citet{VirtualHome} \\
\upd{ThreeDWorld}
& Manip.
& C
& S 
& I,O 
& R,H,SO 
& A,C,D,SS
& \cmark 
& Audio, Physics Interaction, VR
& \citet{TDW} \\
\upd{Sapien}
& Manip.
& C
& S 
& I 
& O 
& C,D,SN,SS 
& 
& Low-level articulation
& \citet{SAPIEN} \\
 \bottomrule
 \end{tabular}
 \label{Tab:SimTask}
}
\end{table}

\subsubsection{Datasets}
\label{sssec:datasets}

\evlp~datasets vary across three main dimensions: visual observations, natural language prompts, and navigation demonstrations. Visual observations, in general, consist of RGB images often paired with depth data or semantic masks. These observations can represent both indoor and outdoor environments from both, photo-realistic or synthetic-based settings. In contrast, language varies in the type of prompt. Language prompts may come in the form of questions \parencite{EQA, eqa_matterport}, step-by-step instructions \parencite{R2R}, or ambiguous instructions that require some type of clarification through dialog or description \parencite{nguyen2019hanna, CVDN}. Language can also vary in terms of complexity of language sequences and scope of vocabulary. Finally, navigation traces differ in aspects like the granularity (or discretization) of the action-space and the implicit alignment that a provided action sequence (or trajectory) has with the other two dimensions. 

\para{VLN Datasets.} Both indoor \parencite{R2R, R4R, BabyWalk,hong2020subinstruction} and outdoor \parencite{TouchDown, mehta2020retouchdown, StreetNav} datasets have been designed for the VLN task. \textcite{R2R} introduced Room2Room (R2R), the first and most widely known indoor VLN dataset. A number of R2R variants have been developed since its introduction. Some of these variants propose longer and more complex navigation sequences. For instance, R4R \parencite{R4R}, R6R, and R8R \parencite{BabyWalk} are formed by concatenating sequences of two, three and four paths from the R2R dataset, respectively. Others aim to provide better alignment between instructions an visual observations. For instance, Fine-Grained Room-to-Room (FGR2R) \parencite{hong2020subinstruction}, proposes finer alignment between instructions and visual inputs. To do so, the authors provide additional supervision for the agent by dividing R2R instructions into sub-instructions and matching them with visual observations along a the corresponding path. Room-Across-Room (RxR) \parencite{ku2020roomacrossroom} uses the PanGEA annotation tool \parencite{PanGEA} to also tackle path-instruction alignment by densely recording and matching 3D pose traces with audio-based instructions. 

Multilingual datasets also exist in VLN. Two such examples of this type of dataset are Cross-Lingual Room-to-Room (XL-R2R) \citep{yan2020monolingual} with instructions in Chinese in addition to English, and RxR \parencite{ku2020roomacrossroom} which uses instructions in Hindi, Telugu, and English. 

The preceding datasets are based on Matterport3DSim where, as mentioned in Section~\ref{ssec:simulators}, navigation is constrained to an undirected graph. Nonetheless, making the underlying navigation space more realistic is another direction into which R2R has evolved. \textcite{VLN-CE} proposed VLN in Continuous Environments (VLN-CE), which transfers the graph-based trajectories of R2R to an unconstrained discrete state-space in AI Habitat. Furthermore, \textcite{RoboVLN} proposed RoboVLN, a dataset which extends the high-level discrete actions of VLN-CE by adding low-level ones that represent the agent's linear and angular velocities between high-level actions. Beyond continuity,there are other short comings in existing datasets that hamper realness. These include biases in path distribution \cite{ku2020roomacrossroom}. Some datasets, such as R2R, have limited variability in path length. In addition they only contains paths that go directly from the origin to a target location. Both these biases impact generalization to new environments and the agent's ability to follow instructions, favoring goal finding rather than instruction following.

In addition to R2R and its variants, other datasets explore outdoor navigation settings. For instance, LANI \parencite{misra2018mapping} is a dataset designed for outdoor navigation in a synthetic environment. LANI evaluates an agent's ability to follow instructions referring to multiple landmarks. \textcite{StreetNav} propose the StreetNav Suite which leverages the StreetLearn environment \parencite{mirowski2019learning} built using RGB panoramas from Google Street View. Similar to R2R, in this dataset navigation is based on an undirected graph but unlike the previous datasets it uses synthetic instructions rather than human-generated ones. 

\para{EQA Datasets.} Datasets in EQA mainly differ based on the type of environment used (Section~\ref{ssec:simulators}) and the type of questions asked. For instance, IQUADv1 \parencite{IQA} is an indoor-based dataset that contains three types of questions: existence questions (\textit{Is there an} \texttt{[OBJECT]} \textit{in} \texttt{[PLACE]}\textit{?}), counting questions (\textit{How many} \texttt{[OBJECT\_1]} \textit{are on the} \texttt{[OBJECT\_2]}\textit{?}), and spatial relationship questions (\textit{Are there} \texttt{[OBJECT\_1]} \textit{in the} \texttt{[OBJECT\_2]}\textit{?}). These questions along with their corresponding answers are generated automatically using templates.

EQAv1 \parencite{EQA} also implements template-generated questions and answers. The type of questions involve scene recognition (\textit{What} \texttt{[ROOM]} \textit{is the} \texttt{[OBJECT]} \textit{located in?}), color recognition (\textit{What color is the} \texttt{[OBJECT]}\textit{?}), and spatial reasoning (\textit{What is} \texttt{[on/above/below/next-to]} \textit{the} \texttt{[OBJECT]} \textit{in the} \texttt{[ROOM]}\textit{?}). The object in a question is single-target, and the possible answer options are room names, color names, and object names. MP3D-EQA \parencite{eqa_matterport}is the extension of EQAv1 to photo-realistic environments. However, it only defines questions about scene and color recognition. Finally, MT-EQA \parencite{MultiAgentEQA} proposes a generalization of EQA where the types of questions involve multiple target objects and rooms. They propose 6 types of questions which compare attribute properties (color, size and distance) between multiple targets (objects and rooms). For instance, \textit{Is} \texttt{[ROOM\_1] [bigger/smaller]} \textit{than} \texttt{[ROOM\_2]}\textit{?}, or, \textit{Does} \texttt{[OBJECT\_1]} \textit{in} \texttt{[ROOM\_1]} \textit{share the same color as} \texttt{[OBJECT\_2]} \textit{in} \texttt{[ROOM\_2]}\textit{?}.

\para{VDN Datasets.} Current indoor-based VDN datasets were built on the R2R dataset. JustAsk \parencite{JustAsk} uses R2R and extends the VLN task by adding an interactivity module where a confused agent can ask an oracle to resolve any ambiguities. In turn, the oracle provides an action to take with some probability of being incorrect. In HANNA \parencite{nguyen2019hanna}, an agent receives traditional VLN instructions and can also request assistance when it is lost or confused. Here, in contrast to JustAsk, an oracle named ANNA provides the agent with a sub-task to help correct its path. As such, the dataset builds a route system simulating human assistance (sub-tasks) on top of the R2R dataset.

\textcite{CVDN} study dialog interactions between agents and human assistants. To do so, they designed the Cooperative Vision-Dialog Navigation (CVDN) dataset which consists of ambiguous and underspecified instructions along with human-based dialog guidance between a navigator and an oracle. Compared to the step-by-step instructions in R2R, their instructions do not initially specify a goal location and require clarification. Thus, resolution is done through the use of dialog between a navigator, who is given the instruction, and an oracle, who knows how to resolve ambiguities. 

Similarly, Talk the Walk \parencite{devries2018talk} consists of a \textit{tourist} and a \textit{guide} that interact with each other using dialog to get the tourist to a target location. The dataset consists of RGB panoramic images of New York City which are used to build a 2D grid-like environment in which the tourist moves. The dialogs for this dataset were crowd-sourced and correspond to trajectories followed from randomly sampled start and target locations. 

\para{EOR Datasets.} There are two datasets in EOR: REVERIE \parencite{qi2020reverie} and Touchdown \parencite{TouchDown, mehta2020retouchdown}. REVERIE is an indoor-based dataset built upon Matterport3DSim. It differs from the R2R dataset in two main aspects: (1) instructions are intended to be more realistic, and (2) the dataset contains object-level annotations. In R2R, instructions are often long and provide step-by-step guidance toward a goal location. \textcite{qi2020reverie} argue that it is unrealistic for humans to give such instructions to a robot. Thus, they propose a dataset containing simplified instructions such as \textit{Bring me a mug}. Additionally, the REVERIE dataset includes object annotations in the form of 2D bounding boxes since the task requires the agent to localize the target object to which the instruction refers (e.g., the mug). 

Touchdown is set in an urban environment. The dataset leverages RGB panoramic images of New York City in Google Street View. It has recently been added to StreetLearn \parencite{mehta2020retouchdown} and made publicly available to the research community. Like StreetNav, navigation is based on an undirected graph. The goal of the agent is to find the location of a hidden teddy bear named Touchdown. Touchdown only appears if the correct location was selected. Compared to REVERIE, the instructions in Touchdown represent a step-by-step guidance toward the goal, similar to VLN. Moreover, these instructions clearly specify where to find the teddy bear, whereas in REVERIE, the agent needs to infer both the object of interest and its probable location in the environment.

\para{EGM Datasets.} All of the contemporary EGM datasets use interactive synthetic environments (Subsection~\ref{ssec:simulators}). EGM datasets are crowd-sourced and validated by human annotators which interact with these environments to verify the alignment between instructions, visual inputs, and interactions. 

The CHAI dataset \parencite{misra2018mapping} is based on household instructions (e.g., \textit{Go to the kitchen and put the cup on the sink}). A single instruction can be divided into sub-tasks requiring multiple navigation and manipulation steps. ALFRED \parencite{ALFRED} is also based on household instructions with multiple goals. The main differences between ALFRED and CHAI are that ALFRED includes expert demonstrations and allows the agent to perform several manipulation actions (e.g., \textit{pick up, turn on, open, etc.}), whereas CHAI only uses an \textit{interact} action.  

ArraMon \parencite{ArraMon} is set on an outdoor environment and consists of two types of instructions: navigation and assembly. Navigation instructions provide step-by-step guidance toward a goal object which the agent requires to find and collect. Assembly instructions specify where does the agent need to place each the collected item. These directives come in turns, i.e., a navigation instruction is followed by an assembly one. Each episode is defined by two turns. Lastly, CerealBar \parencite{CerealBar} also set outdoors, considers a turn-based game between two agents, a \textit{leader}, and a \textit{follower}. The leader gives step-by-step instructions to collect game cards of different colors, shapes and count, and the follower has to execute them. 

\begin{table*}[ht!]
\rowcolors{2}{gray!10}{white}
\caption{\small \ann{Reviewer C, ``Minor Comments", item \#2} Summary of \evlp~dataset statistics. \textit{MP3D} refers to the Matterport3D simulator \citep{Matterport3D}. \textit{CS} refers to crowd-sourced instructions and \textit{G} refers to generated instructions. We use \textit{EN} for English, \textit{CH} for Chinese, \textit{HI} for Hindi and \textit{TE} for Telugu. Then, we use \textit{N} for navigation and \textit{M} for manipulation. \rev{The vocabulary sizes provided are taken from the original works and account for all unique words generated using that work's tokenizer.} Finally, missing fields, indicated by hyphens (-), were not computed by the originating works.}
\label{tbl:dataset}
\centering
\small

\resizebox{\textwidth}{!}{\begin{tabular}{lcccccccccccc} 
\toprule
& & & \multicolumn{3}{c}{Navigation Information} & &\multicolumn{5}{c}{Language Information}\\ 
\cmidrule{4-6} \cmidrule{8-12} 

\textbf{Dataset}   & \textbf{Task}  & \textbf{Simulator}   & \begin{tabular}[c]{@{}c@{}} \textbf{Avg.}\\ \textbf{\# Steps}\end{tabular}   & \textbf{\# Trajectories}  & \textbf{Navigation} & & \begin{tabular}[c]{@{}c@{}} \textbf{\# Instr. / Dialog /}\\ \textbf{Questions}\end{tabular} & \begin{tabular}[c]{@{}c@{}} \textbf{Avg. Words Per}\\ \textbf{Instruction}\end{tabular} & \begin{tabular}[c]{@{}c@{}} \textbf{Instruction}\\ \textbf{Type}\end{tabular}  & \begin{tabular}[c]{@{}c@{}} \textbf{Vocabulary}\\ \textbf{Size}\end{tabular} & \textbf{Language} & \textbf{Reference}\\ \hline

R2R  & VLN & MP3D  & 5 & 7,189 & graph & & 21,567 & 29 & CS & 3,100 & EN & \textcite{R2R}\\ 

R4R  & VLN & MP3D & 20.5 & 7,189 & graph & & 279,810 & 58.4 & CS & 3,100 & EN & \textcite{R4R}\\ 

R6R  & VLN & MP3D  &  - &  - & graph & & 12,5409 & 91.2 & CS & 3,100 & EN & \textcite{BabyWalk}\\ 

R8R  & VLN & MP3D  & - & -& graph & & 138,004 & 121.6 & CS & 3,100 & EN & \textcite{BabyWalk}\\

FGR2R & VLN & MP3D  & - & 7,189  & graph & & 21,567 & 29 & CS & 3,100 & EN & \textcite{FineGrainedR2R}\\

XL-R2R  & VLN & MP3D  & - & 5,798  & graph & & 17,394 & - & CS & EN: 1,583, CH: 1,134 & EN, CH & \textcite{yan2020monolingual}\\

RXR  & VLN & MP3D  & 7 & 16,522  & graph & & 126,000 & EN: 101, HI: 76, TE: 56 & CS & - & EN, HI, TE & \textcite{ku2020roomacrossroom}\\

VLN-CE  & VLN & Habitat  & 55.8 & 4,475  &  discrete & & 13,425 & - & CS & - & EN & \textcite{VLN-CE}\\

RoboVLN & VLN & Habitat & 326 & 3,177  & continuous & & 9,533 & - & CS & - & EN & \textcite{RoboVLN}\\

LANI  & VLN & Unity3D & 24.6 & 6,000 & grid & & 28,204 & 12.1 & CS & 2,292 & EN & \textcite{misra2018mapping}\\

StreetNav & VLN & StreetLearn  & 125 & 613,000 & graph & & $\sim$2.5M & 7 & G & - & EN & \textcite{StreetNav}\\

IQAv1 & EQA & AI2-THOR & -  & - & discrete & & 75,000 & - & G & 70 & EN & \textcite{IQA}\\

EQAv1  & EQA & House3D  & - & - & discrete & & 5,000 & - & G & - & EN & \textcite{EQA}\\ 

MT-EQA  & EQA & House3D  & - & - & discrete & & 19,287 & - & G & - & EN & \textcite{MultiAgentEQA}\\ 

MP3D-EQA  & EQA & Habitat & - & - & discrete & & $\sim$1,100 & - & G & - & EN & \textcite{eqa_matterport}\\ 


CVDN & VDHN & MP3D  & 17.4 & 7,000 & graph & & 2,050 & 81.6 & CS & $\sim$1,100 & EN & \textcite{CVDN} \\

HANNA & VDHN & MP3D  & - & - & graph & & 21,594 & - & CS & 2,332 & EN & \textcite{nguyen2019hanna} \\

Talk the walk  & VDHN & Talk the Walk  & 62 & - & grid & & 10,000 & - &  CS & 10,000+ & EN & \textcite{devries2018talk}\\

REVERIE  & EOR & MP3D  & - & - & graph & & 21,702 & 18 & CS & 1,600 & EN & \textcite{qi2020reverie}\\

(Re)Touchdown  &  EOR & StreetLearn  & - & - & graph & & 9,326 & 108 & CS & 5,625 & EN & \textcite{TouchDown, mehta2020retouchdown}\\

CHAI  & EGM & Chalet  & 54.5 & - & discrete & & 13,729 & 10.1 & CS & 1,018 & EN & \textcite{misra2018mapping}\\

ALFRED  & EGM & AI2-THOR  & 50 & 8,055 & graph & & 25,726 & 50 & CS & $\sim$100 & EN & \textcite{ALFRED}\\

ArraMon  & EGM & ArraMon  & 75.78 (N), 13.68 (M) & 7,692 & discrete/grid & & 30,768 & 48 (N), 21 (M) & CS & - & EN, HI & \textcite{ArraMon}\\

CerealBar  & EGM & CerealBar & 8.5 & 1,202 & grid & &  23,979 & 14 & CS & 3,641 & EN & \textcite{CerealBar}\\ 

\bottomrule
\end{tabular}}
\end{table*}

\section{Open Challenges in \EVLP} \label{sec:openchallenges}


\ann{Reviewer C, ``Organization", item \#1}
\subsection{\rev{New Directions in EVLP Research}}
\label{subsec:challenges}

\rev{Research in \EVLP~has experienced a quick rise in popularity, due to recent advances from the robotics, computer vision, and natural language processing communities on vision-language grounding and situated learning. As the \evlp~research field evolves, we highlight three promising directions, in the pursuit of more ubiquitous human-robot interaction and better agent generalisation. Firstly, we describe \textit{social interaction} (Section \ref{subsec:social_interaction}) as a necessary progression from static dialogue contexts to more dynamic ones, where new collaborative and assistive capabilities of agents can more effectively emerge. Next, we discuss the need for developing agents in \textit{dynamic environments} (Section \ref{subsec:social_interaction}), which encourage agents to incorporate reasoning strategies that are robust to environment uncertainty and non-stationarity. Finally, we discuss a vision for \textit{cross-task robot learning} (Section \ref{subsec:unified_robot_learning}) for cross-task transfer, wherein agents may acquire experience from related modality-centric tasks, before deployment to shared multimodal settings with significant task overlap.}
\upd{\subsubsection{Social Interaction}}
\label{subsec:social_interaction}
\upd{Social agent interaction is born out of a desire for an agent to carry out tasks and make decisions collaboratively with other agents in a complex, multimodal world. Particularly compelling application formats are those that require the intelligent system to perform \evlp~tasks, wherein agents not only perceive the world to extract knowledge from it but also actuate it, to create change that is beneficial to others. Of particular importance is a shared understanding of the physical space and task structure, throughout the interactions with other agents in the environment---a capability that is absent from disembodied dialogue interaction systems. Such elevated understanding gives the agent access to previously-inaccessible interaction scenarios, which can manifest in assistive technologies for those with physical disabilities or allow for the deployment of search-and-rescue systems in which a layperson can ask for help and explain their circumstances.}

\upd{Despite the importance of collaboration for real-world deployment, current \EVLP~tasks (Section \ref{sec:tasks}) tend not to include significant social interaction. Even in the case of the Vision and Dialogue Navigation (VDN) and Embodied Goal-directed Manipulation (EGM) tasks, the dialogue interaction components are \textit{static}, i.e., pre-generated datasets of dialogue histories are provided to the agent, during training and testing. Without tasks and simulators that enable more \textit{dynamic} interaction with agents, it is challenging to develop approaches that manifest behaviour typical of online collaborative settings \parencite{fried2021reference, CerealBar}, such as: (i) modeling other agent(s) conceptual schema of the world, (ii) incorporating direct feedback about task progress, (iii) and generating statements and questions (e.g., for informing other agent(s) about task progress, requesting assistance, resolving ambiguity in received instructions, asking for alternative commands, etc.) \parencite{padmakumar2021teach}.}

\upd{In order for \evlp~agents to perform complex tasks that require dynamic interaction with other agents, the underlying simulation environments and task structures must support the desired interaction settings. Primarily, environments must enable the representation of others' anticipated actions, mental state, and previous behaviour, which has been shown to be critical in related areas, such as social navigation \parencite{tsai2020generative, vemula2018social, mavrogiannis2021core}, natural language processing \parencite{fried2018unified, fried2021reference, zhu2021few}, and human-machine interaction \parencite{newman2018harmonic, jeon2020shared, social_robotics}. We highlight these related fields as inspiration for this new direction in \evlp~research.}
\upd{\subsubsection{Dynamic Environments}}
\label{subsec:dynamic_environments}
\upd{Most contemporary tasks and agents assume that the underlying environments are stationary and unchanging. Instead, we wish to highlight the notion of \textit{dynamic} environments, which we define as those that change due to events both inside and outside of the ego-agent's locus of control. The RobustNav task \parencite{chattopadhyay2021robustnav} extended the RoboTHOR visual navigation environment \parencite{RoboTHOR}, to introduce structured perturbations on perception and transition dynamics, with the intention of encouraging agents to learn robust navigation policies in lieu of these corruptions. These perturbations are static with respect to the agent's task-execution, however, and do not change overtime. Other tasks, such as ALFRED \parencite{ALFRED} and TEACh \parencite{padmakumar2021teach}, introduced entity-based interaction and unrecoverable states. However, these interventions can only be caused by the agent and cannot occur independently.}

\upd{In real-world environments, the state of the environment can change without any intervention from the agent. Whether these changes should result from another agent (cognitive or otherwise) acting on the environment, from physical dynamics of the environment (e.g., a book sliding off a shelf), or from temporal dependencies on available interactions (e.g., interactions with certain objects are only available at certain times of day), the planning and reasoning complexities introduced by dynamic environments are challenging, yet largely unexplored, properties. We advocate for the introduction of these dynamic elements to \evlp~tasks, in order to advance one step closer to enabling agents that accommodate the complexities of real-world deployment.}
\upd{\subsubsection{Cross-Task Robot Learning}} 
\label{subsec:unified_robot_learning}

\evlp~tasks are currently evaluated separately from one another, which fails to capture the underlying skills shared between tasks and the overall progress made as a field. Being able to follow instructions (e.g., as in the VLN task family) and also answer questions about an environment (e.g., as in the EQA task family) are not and should not be treated as mutually-exclusive tasks. Rather than propose and evaluate tasks independently, tasks could be combined according to more unifying requirements on agent capability. \ann{Reviewer C, ``Taxonomy'', item \#2} \rev{Similar notions of cross-task learning gave rise to seminal modeling strategies (e.g., hierarchical task decomposition) and problem definitions (e.g., mobile manipulation, under EGM), which remain quite relevant to this day, despite their classical foundations (see Section \ref{ssec:navigation}). Combinations of existing task families may, likewise, yield new insights and advances.}

Recent works have begun to test over several environments, but often within the same task family \parencite{PREVALENT}. Extending these types of benchmarks could allow for the use of transfer learning to bootstrap models. Moreover, it would allow for shared metrics that measure underlying skills of a particular agent. The idea that pre-trained models have specific abilities required for downstream tasks is not new and has been explored by the NLP \parencite{olmpics} and robotics communities \parencite{he2018zero}. Extending this to \evlp~tasks could prove beneficial to evaluating progress in the field as a whole rather than over very specific benchmarks. Zero-shot learning between tasks, with the idea that a model which can ground language should be as capable at manipulation as it is at navigation could be used to this end. Otherwise, specific benchmarks could also be designed to test for specific abilities, such as the ability to ground left and right, ability to recognize if an object is invalid, ability to recognize objects. In this vein, \textcite{zhu2021diagnosingVLN} test the impact of masking direction and object information in text on performance to evaluate how current VLN models use these types of information. They find that the type of information that led to the biggest decrease, either object or directional information, was dataset dependent. This finding means that shifts in information type in instruction could impact model performance. 

Another test was conducted to evaluate the ability of models to ground text to vision by masking objects in the images found in the instructions. Doing so resulted in a limited decrease in performance, casting doubt on the ability of current models to ground. More works in this line could help understand and evaluate potential weaknesses in current models and how to improve their ability to generalize. We provide further discussion on the need for improved cross-task agent generalizability in Section \ref{subsubsec:evaluationparadigms}.

\subsection{Use of Domain Knowledge}
\label{ssec:domain_knowledge}

Using domain knowledge to guide the learning progress of models has seen a recent resurgence in various domains, such as: autonomous driving \parencite{park2020diverse, chen2019learning, li2020evolvegraph}, autonomous racing \parencite{herman2021learn, chen2021safe, francis2022l2rarcv}, robot navigation \parencite{das2018neural, VSNScenePriors}, natural language processing \parencite{andreas2016neuralmodulenetworks, ravichander2017would, ma2019generalizable, ma2020zeroshotqa, oltramari2020neurosymbolic, li2020lexicallyconstrained, lu2020neurologic}, sensing and control \parencite{8431085, francis2019occutherm, chen2020cohort, munir2017real}, and many others. Indeed, domain knowledge comes in many forms, such as in graphical models, physics-based constraints, admissibility conditions, auxiliary objectives, knowledge distillation, pre-training steps, symbolic commonsense knowledge, and many others. While domain knowledge holds the promise of improving agents' sample-efficiency, interpretability, safety, and generalizability, the challenge exists in how to effectively express and utilize the domain knowledge in an arbitrary problem. In this section, we highlight \textit{pre-training} and \textit{commonsense knowledge}, in particular, as these are two manifestations of domain knowledge that show incredible promise for endowing agents with the aforementioned attributes.

\subsubsection{Pre-Training} A common strategy for injecting agents with external domain knowledge is to borrow information from other tasks, models, and domains through generalized pre-training steps. Stemming from principles of non-convex optimization, the idea behind pre-training is that task-specific models and agents can start with a more optimal parameterization for a given task (for subsequent training or ``fine-tuning"), after having already undergone partial training on other similar tasks. However, the selection of these pre-training tasks is not to be taken lightly: pre-training models on tasks that contain counterfactual or causally-confusing samples, with respect to those in the intended downstream task, can cause models to get stuck in sub-optimal local minima during the fine-tuning process. In response to these challenges, pre-training tasks have been carefully designed and coupled with popular high-capacity models, in such domains as image classification \parencite{Resnet, MultiModalSurvey} and natural language processing \parencite{devlin2018bert, yang2019xlnet, ma2020zeroshotqa}, in attempts maximize the generalizability of transferred or fine-tuned approaches. While there is some progress in the context of specific multimodal problems \parencite{VLNBERT, PREVALENT, lu2019vilbert}, where the cross-modal reasoning is performed primarily at a single time-step, challenges remain for developing generalizable pre-training strategies that encompass the scope of the broader \evlp~task family, wherein cross-modal reasoning must occur over arbitrary temporal extents and alongside complex transition dynamics.

\subsubsection{Commonsense Knowledge} In the real world, much of the background information that governs the interactions amongst objects, cognitive agents, and the environment is \textit{implicit}|that is to say that this knowledge is learned (by humans) through experience and is not commonly expressed explicitly, as we undertake our actions in the world. Tasks that require this tacit or ``commonsense" knowledge\footnote{Common sense is broad and inherits several concepts from related domains, such as cognitive science and psychology. For a brief treatment of the various types of commonsense knowledge, in the context of neural prediction on AI tasks such as question answering in NLP, we refer interested readers to the works by \textcite{ma2019generalizable, ma2020zeroshotqa}.} are notoriously difficult for machines; even for tasks, in which models have recently enjoyed substantial performance improvements by other means, researchers find that commonsense knowledge still represents a fundamental cognitive gap in model- versus human-level performance, in scene understanding capability \parencite{ma2019generalizable, ma2020zeroshotqa, oltramari2020neurosymbolic}. Challenges arise, however, if the \textit{type} of commonsense knowledge to be used (e.g., declarative, relational, procedural, sentiment, metaphorical, etc.) is not chosen to align well with the representation of that knowledge or with the implicit semantic characteristics of the downstream task. 

Commonsense knowledge acquisition and injection in models remains an active research area in NLP~\parencite{talmor2018commonsenseqa,ma2019generalizable,ma2020zeroshotqa, li2020lexicallyconstrained}, with some works proposing to ground observations with structured commonsense knowledge bases, directly, thereby improving downstream performance on relevant tasks. However, the use of commonsense knowledge, in the context of \evlp~tasks remains largely unexplored. 
As the ultimate goal of \evlp~tasks is to develop intelligent agents that are capable of solving real-world problems in realistic environments, it is reasonable to consider providing models with structured external knowledge from the world \parencite{VSNScenePriors}. In general, \evlp~tasks can be viewed as a series of tasks that test agents' commonsense reasoning, where the agent is required to learn general skills that can be transferred to unseen contexts. 

\subsection{Agent Training Objectives}
\label{sssec:openchall:obj}

Selecting the appropriate training objective(s) for agents undertaking a given task has been a long-standing problem in machine learning and artificial intelligence; this selection depends on the nature of the available training signal(s) (reward/cost, full supervision, limited or non-existent supervision) and on the degree to which external knowledge (e.g., auxiliary objectives, constraints) is deemed necessary for effectively biasing agent behaviour. For \evlp~tasks, specifically, the selection of training objectives is made more challenging by the complex nature of the environments, often necessitating frameworks that consist of more than one biasing strategy. For example, many state-of-the-art models in VLN use self-supervised learning or multi-task learning in a pre-training stage then supervised learning in a fine-tuning stage, in order to borrow information from large external datasets \parencite{VLNBERT}. Compelling approaches in the EQA task family often start with supervised (or imitation) learning to first learn from available expert traces, then use reinforcement learning in a fine-tuning stage, with the hope of finding a more generalisable model parameterisation for unseen environments \parencite{EQA, eqa_matterport}. Of the popular learning paradigms in use, for a given task, it is challenging to define which permutation(s) yield the best downstream and cross-domain results. However, given the underlying motivation of optimising for generalisability and interpretability, explicit treatment should be given to finding the learning paradigm(s) that most effectively integrate information for various related sources and generalises agents' inductive biases to new environments; the training paradigms should include explicit mechanisms (constraints, regularisation strategies) for encouraging these properties we hope to imbue.

\subsection{Model Evaluation Paradigms} 
\label{subsubsec:evaluationparadigms}

\noindent Datasets and simulation environments are the primary driving forces behind \evlp~research, since one's ability to measure model efficacy relies on the availability of strong testing scenarios and the appropriate evaluation criteria. Current \evlp~tasks are implemented as a set of goals and metrics, atop pre-existing simulators or datasets. In this section, we urge the community to consider and prioritise the deployment of \evlp~agents to real-world settings. As examples of potential task applications, we may imagine VLN models to be used for object-delivery and retrieval, EQA models for search and reconnaissance scenarios, VDN models for shared autonomy and human-machine collaboration, EOR models for scene understanding and surveillance, and EGM models for such applications as household robotics and manufacturing. Here, task metrics can have conflicting objectives, where they need to resemble measures of what ``good" agent behaviour might look like in the real-world, despite being subject to the practical limitations of the dataset or simulator itself (e.g., physical fidelity, semantic saliency, phenomenological coverage). Specifically, we assert that various \evlp~tasks and metrics may be improved on the basis of three dimensions: (i) simulator realness, (ii) dataset realness, and (iii) tests for model generalisability. 

\subsubsection{Simulator Realness} 

Simulation-based training is intended to serve as a proxy for training agents in the real-world, particularly in situations where operating or fully-supervising real-world agents would be prohibitively expensive or impossible. Furthermore, simulation-based training and execution are especially attractive when modeling a sequential learning problem (i.e., in which the environment's state and an agent's observations are functions of the agent’s previous action), since offline datasets do not allow for such recursive interaction with an environment. There are limitations, however, in how effectively scientists and practitioners can encourage the desired model behaviour to emerge for real-world use-cases---chief among these limitations are those of the simulation environment itself as well as discrepancies in how task metrics are defined. \ann{Reviewer C, ``Claims'', items \#2 and \#3} \rev{Compared to the real-world, simulators have reduced physical fidelity (e.g., discretized navigation graphs, synthetic visual contexts) and limited semantic/phenomenological coverage (e.g., time-invariant weather and obstacles). As a result, a number of real-world physical challenges in robotics and computer vision (e.g., visual affordance learning, proprioceptive control, system identification) are abstracted away in simulated EVLP tasks. While this may allow for a concerted effort for pursuing other aspects of advanced reasoning in agents, agents trained in these environments risk overfitting to the incomplete transition dynamics provided by the simulated environment, which are only approximations of the \textit{true} dynamics that govern corresponding natural generative processes \parencite{alami1998architecture}. Furthermore, the task execution primitives (e.g., \texttt{pick}, \texttt{place}, \texttt{move}), which have become commonplace in simulated EVLP tasks as abstracted actions, may not always translate to the same abstractions in the real-world, e.g., due to noisy execution \parencite{chattopadhyay2021robustnav}.} \upd{Because this dissonance reduces models' immediate viability for real-world deployment, we assert the importance of increased attention from the computer vision and robotics communities on the topics of simulation-to-real transfer, unseen generalisation, out-of-distribution prediction, and domain adaptation.} In situations where task metrics are inappropriately defined (e.g., only end-goal success rate metrics are available, ignoring collisions and sharp movements), trained models are optimised according to incomplete objectives, which serve only to distinguish between entries on AI task leaderboards rather than fundamentally assess model readiness for real-world deployment. \upd{We encourage the definition of metrics that assess intermediate agent behaviours and task efficiency, as opposed to simply indicating task completion.}

\subsubsection{Dataset Realness} Dataset-based agent training thrives in many situations, including those that dictate modeling the space of all observations and agent actions as probability distributions (as in normalising flow-based approaches, variational Bayesian methods, etc.), when performing knowledge distillation with privileged information (as in learning from demonstrations), when warm-starting models before policy refinement (e.g., pre-training, data augmentation), or when introducing auxiliary objectives to the training procedure (e.g., learning skills, supervising stopping criteria).\footnote{One may reasonably argue in favour of dispensing with the datasets and, instead, using a simulator (should one be available). However, we offer a reminder that using labeled datasets constitutes performing knowledge distillation with privileged information~\parencite{lopezpaz2016unifying}, as in Imitation Learning (IL), which has provably faster convergence time and reduced model capacity requirements, compared to learning \textit{without} privileged information. The challenges facing IL ~\parencite{ross2011reduction, dehaan2019causal} result primarily from the lack of comprehensiveness and robustness in the expert (teacher) traces, not in the notion of using an expert in itself. Furthermore, IL approaches have become the \textit{de facto} alternative to formulating \evlp~tasks as POMDPs with sparse rewards.} Despite these momentary advantages over simulation, physical fidelity and semantic coverage requirements are even more pronounced in dataset-based training, since, now, limited (or no) observation of the environment's transition dynamics is available to agents as additional supervision. Indeed, rare-event coverage is particularly challenging for models in dataset-based training settings, due to the imbalanced support that leads to unrealistic priors and ill-prepared posteriors. Another limiting factor in the use of datasets for real-world model deployment is the adoption of unrealistic assumptions about the design or collection of data, e.g., the inclusion of programmatically-generated template instructions that would likely not be observed in the real-world. Next, the modalities provided by various datasets reflect only a subset of the sensory information that a cognitive agent may utilise for solving \evlp~tasks in the real world. For example, few datasets have the notion of ambient environmental sounds (e.g., running sink faucets, vibrations from home appliances), despite this information being crucial to how humans interact with the world~\parencite{chen2020soundspaces}. Moreover, dataset metrics vary in the agent behaviours they seek to encourage, i.e., by way of the model training objectives they necessitate (see Section \ref{sssec:openchall:obj}). We want agent behaviour to manifest, free from the spurious variation in the visual observations, natural language, and expert traces; we want agent behaviour to remain invariant to irrelevant features (e.g., environment-specific scene backgrounds, linguistic synonyms and variance in sentence structure, etc.) and, accordingly, we encourage the emergence of \evlp~datasets that enable agents to be simulator-agnostic. The same metrics may be defined across \textit{multiple} environments, enabling ``purer" skill affordances. Finally, special care should be given to the formats in which datasets are released and reported: characterising (i) the action space, (ii) instruction length (average and range), (iii) vocabulary size, (iv) data collection procedure, and (v) data availability (e.g., open-source or academic use permitted).

\subsubsection{Tests for Generalisability} \evlp~datasets, such as those listed in Table \ref{tbl:dataset} of this paper, have been conventionally separated across domain and task lines \citep{StreetNav, TouchDown, R2R, EQA}. In the context of real-world deployment, however, models are expected to encompass such properties as transferability to unseen environments and robustness to model and environmental uncertainty. Agents should be able to perform the same \evlp~task, regardless of the nuances in the visual background; agents should be capable of driving in both urban and highway scenarios; agents trained on a relevant set of tasks should enjoy an advantage on similar unseen tasks. We assert that evaluation paradigms should assess agents using explicit tests for: generalisability across domains; generalisability across tasks. The need for these additional evaluation assessments is not new. As the ubiquity of deployed approaches increased while the challenges in robust multimodal alignment continued to abound, other communities, such as visual question answering (VQA), have faced similar changes in requirements. \textcite{agrawal2018dont} produced the VQA-CP dataset, an extension of the original VQA task \parencite{VQA} which contained multiple validation splits from different environments. While, in \evlp~tasks, there is already the notion of seen and unseen environment splits \parencite{R2R, EQA, IQA, ALFRED}, the multiple dimensions of possible variation in the unseen split (distribution of objects, distribution of backgrounds, distribution of indoor layouts) remain largely unstudied; there are no guarantees that these splits are representative of the types of desired real-world scenarios. Having a test bed with different test splits would be beneficial, as it would allow researchers to verify if their models are learning spurious (or non-transferable) correlations in the observations or if the models are actually learning to plan.
The authors also advocate for the assessment of agents' abilities to generalise to longer path lengths and to other tasks. The \evlp~community already has several datasets that would facilitate this analysis, which broadly fall into two general categories: path concatenation and path decomposition. Path concatenation, such as R4R, R6R, R8R \parencite{R4R, BabyWalk} works by joining paths that start and terminate near one another to generate longer paths. Agents can be trained on these longer paths or evaluated over them \parencite{R4R}. Rather than building longer paths from the same dataset, path decomposition breaks down the path into fine-grained instructions \parencite{FineGrainedR2R, BabyWalk}. The agent is trained over those and then evaluated over the larger dataset with longer instructions. Note that path concatenation and decomposition are not mutually exclusive \parencite{BabyWalk}. Extending these techniques to other datasets or developing new techniques would be one way of evaluating how length and complexity impact performance.

\section{Conclusion}

As a field in development, \evlp~has never been fully documented. In this paper we proposed a taxonomy which covers: tasks in \evlp, learning paradigms and training techniques used, evaluations, open challenges. We provided a framework to discuss existing and future tasks based on the skills required to solve them. We discussed usage of training paradigms such as reinforcement learning and supervised learning, common training tricks such as data augmentation, and commonly used loss functions. Evaluation of models and a framework to discuss what aspects of performance each of these metrics covers. Finally the challenges currently being tackled in the field and what challenges still remain to be tackled. Specifically we discuss issues that could prevent real world deployment, such as a lack of generalization, robustness, simulator realness, and lack of interactions. With the speed of improvements in the field we feel that these challenges could be tackled and allow for real world deployment of these \evlp~agents.

\section*{Acknowledgements}

\upd{The authors thank Alessandro Oltramari, Yonatan Bisk, Eric Nyberg, and Louis-Philippe Morency for insightful discussions; we thank Mayank Mali for support throughout the editing process}, and we thank the JAIR reviewers for their valuable feedback. This work was supported, in part, by a doctoral research fellowship from Bosch Research, by the U.S. Air Force Office of Scientific Research, under award number FA2386-17-1-4660, and by Jonathan's dual affiliation with the Human-Machine Collaboration group at Bosch Research Pittsburgh. The views expressed in this article do not necessarily represent those of the aforementioned entities. All authors contributed equally to the manuscript; author list in alphabetical order. Professor Jean Oh is the corresponding author, reachable at \textit{jeanoh@cmu.edu}.

\bibliography{reference}
\bibliographystyle{apalike}

\end{document}